# Transfer Prototype-based Fuzzy Clustering

Zhaohong Deng, *IEEE Senior Member*, Yizhang Jiang, *IEEE Member*, Fu-Lai Chung,
Hisao Ishibuchi, *IEEE Fellow*, Kup-Sze Choi, *IEEE Member*, Shitong Wang

***Abstract*—Traditional prototype-based clustering methods, such as the well-known fuzzy c-mean (FCM) algorithm, usually need sufficient data to find a good clustering partition. If available data are limited or scarce, most of them are no longer effective. While the data for the current clustering task may be scarce, there is usually some useful knowledge available in the related scenes/domains. In this study, the concept of transfer learning is applied to prototype-based fuzzy clustering (PFC). Specifically, the idea of leveraging knowledge from the source domain is exploited to develop a set of transfer prototype-based fuzzy clustering (TPFC) algorithms. First, two representative prototype-based fuzzy clustering algorithms, namely, FCM, and fuzzy subspace clustering (FSC), have been chosen to incorporate with knowledge leveraging mechanisms to develop the corresponding transfer clustering algorithms based on an assumption that there are the same number of clusters between the target domain (current scene) and the source domain (related scene). Furthermore, two extended versions are also proposed to implement the transfer learning for the situation that there are different numbers of clusters between two domains. The novel objective functions are proposed to integrate the knowledge from the source domain with the data in the target domain for the clustering in the target domain. The proposed algorithms have been validated on different synthetic and real-world datasets. Experimental results demonstrate their effectiveness in comparison with both the original prototype-based fuzzy clustering algorithms and the related clustering algorithms like multi-task clustering and co-clustering.**

## I. INTRODUCTION

Transfer learning [1] is receiving more and more attentions in the fields of machine learning and data mining. It is based on the assumption that some useful information is available from a reference scene while the data in the current scene are scarce for the current learning task. The current scene and the reference scene are commonly named as the target domain and the source domain, respectively. Currently, transfer learning has been demonstrated as a promising approach to obtain an effective model for the target domain by effectively leveraging useful information from the source domain in the learning process. Its main assumptions or characteristics can be summarized as follows. First, the data in the target domain is not sufficient to generate a good model for the target domain. Second, the source domain is assumed to be related to the target domain in a sense that they are different but similar to some extent, making the trained model on the source domain not directly applicable to the target domain but useful for the learning in the target domain. Third, the modeling effect for the target domain is the main concern while that of the source domain is usually not a focus.

In the past decade or so, transfer learning has been studied extensively for various applications such as text classification [2] and indoor WiFi location estimation [3]. Application fields of transfer leaning in the existing studies can be generally divided into four categories: 1) classification [2-8], 2) feature extraction [9, 10], 3) regression [11-15], and 4) clustering [16, 17]. While the first three tasks have been studied quite extensively, research on transfer clustering is still limited despite of the wide range of real-world clustering applications. Clustering has long been a major research topic and many clustering algorithms have been proposed. They can be classified as: 1) prototype-based clustering [18-23, 46-50]; 2) density-based clustering [24-28]; 3) graph-based clustering [29-32]; and 4) clustering based on other data modeling mechanisms [33-35]. Among these four types of algorithms, prototype-based clustering such as k-means is perhaps the most popular one and in fact has been extensively studied. In this paper, we focus on transfer learning for this type of clustering algorithms and propose several representative transfer prototype-based clustering algorithms in the context of fuzzy clustering.

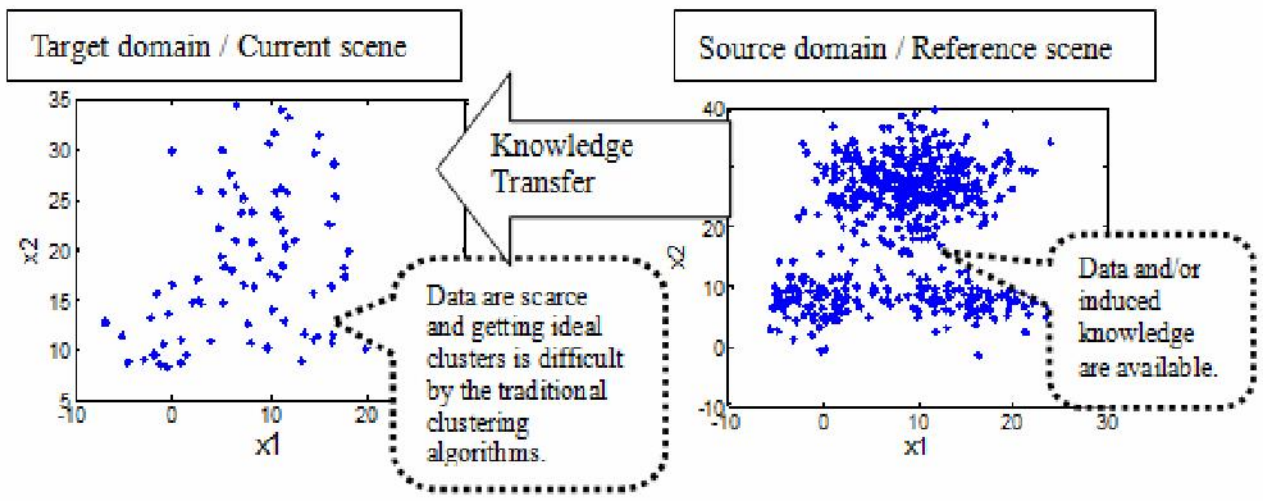

Fig. 1 Illustration of a situation where transfer learning is required for the clustering task.

Fig. 1 illustrates a situation where transfer learning is useful. As shown in Fig.1 (left part), it is difficult to obtain an ideal clustering result from the data in the target domain using the

Manuscript received xxxx. This work was supported in part by the Outstanding Youth Fund of Jiangsu Province (BK20140001), National Natural Science Foundation of China (61170122, 61272210), the Fundamental Research Funds for the Central Universities (JUSRP51321B), the Ministry of education program for New Century Excellent Talents (NCET-120882)，the Hong Kong Polytechnic University (G-UA68, G-UA3W).

Z.H. Deng, Y.Z. Jiang and S.T. Wang are with the School of Digital Media, Jiangnan University, Wuxi 214122, China (e-mail: dzh666828@ aliyun.com; 7121606003@vip.jiangnan.edu.cn; wxwangst@aliyun.com)

F.-L. Chung is with the Department of Computing, Hong Kong Polytechnic University, Hong Kong (e-mail: cskchung@comp.polyu.edu.hk)

H. Ishibuchi is with the Department of Computer Science and Intelligent Systems, Osaka Prefecture University, Osaka 599-8531, Japan (e-mail:hisaoi@cs.osakafu-u.ac.jp).

K.S. Choi is with the School of Nursing, the Hong Kong Polytechnic University(e-mail: kschoi@ieee.org)

traditional prototype-based clustering algorithms. However, if the information from the reference scene, i.e., the source domain, in Fig.1 (right part) is considered, more promising clustering results can be expected. For example, the induced knowledge from the source domain, such as the cluster centers, can be employed to guide the clustering process in the target domain. It can be seen from this example that the key to developing effective transfer clustering algorithms is to realize effective knowledge/information transfer from the source domain to the target domain. It seems that the simplest way to obtain information from the source domain is to merge the data from the two domains and use the merged data in the target domain. However, the direct use of the source domain's data in the target domain may suffer from the following problems. First, important information such as the distribution difference between the source and the target domains will be lost if we simply merge the data from the two domains. Second, the data of the source domain should not be directly used in the target domain. This is because there exist similarities as well as differences in the data distributions between the two domains. This is also because the source domain's data may dominate the distribution of the merged data when the size of the data in the source domain is much larger. Third, due to the necessity of privacy protection in some applications, such as customers' personal information, the raw data of the source domain may not be directly available. All these problems should be properly addressed in order to develop an effective transfer learning strategy for prototype-based clustering methods.

In this paper, transfer learning for prototype-based clustering methods is exploited. While the classical center prototype-based algorithms have been extensively studied and applied, in the past decade the subspace prototype-based algorithms have been attracting more and more attentions of researcher due to their excellent performance on clustering high dimensional data. In view of their representativeness, two prototype-based fuzzy clustering (PFC) algorithms, namely, fuzzy c-mean (FCM) with cluster center prototypes, and fuzzy subspace clustering (FSC) with cluster subspace prototypes, have been chosen to incorporate with knowledge leveraging mechanism to develop the corresponding transfer PFC (TPFC) algorithms. The knowledge induced in the source domain, such as the cluster centers and the subspace of the clustering results, has been appropriately used to boost the clustering performance in the target domain. A novel objective criterion is proposed to integrate the knowledge from the source domain with the data of the target domain for the learning of the cluster prototypes in the target domain. Thus, transferring knowledge of the source domain can effectively make up the deficiency in the target domain's data for clustering.

As mentioned, transfer clustering has not been widely studied. In [16], the self-taught clustering (STC) has been proposed as a very first transfer clustering algorithm based on mutual information. In [17], a transfer learning version of spectral clustering is proposed. These two transfer clustering algorithms are not developed for prototype-based clustering. There exist some obvious differences between these related works and the proposed TPFC algorithms. The self-taught clustering is based on mutual information (MI). Therefore enough data in both source and target domains are assumed available for proper estimation of the probability densities and computation of the MI accordingly [16]. Moreover, it assumes that the raw data of the source domain are available. However, this is not always the case in some applications. In the proposed algorithms, the raw data of the source domain are not required and available data in the target domain can be scarce. The transfer spectral clustering algorithm [17] has been specifically designed for the spectral clustering while our work focuses on the prototype-based fuzzy clustering methods.

Transfer clustering is also related to co-clustering [36], collaborative clustering [37, 38] and multi-task clustering [39, 40], where multiple clustering tasks are usually handled simultaneously and they may cooperate with each other in order to improve the clustering performance of all clustering tasks involved. By comparing co-clustering, collaborative and multi-task clustering with the proposed TPFC algorithms, we can see that these three types of related clustering techniques assume that the multiple clustering tasks involved are equally important and each can benefit from the others in the clustering process. However, the proposed TPFC algorithms only focus on the clustering task in the target domain. The source domain is used only to enhance the clustering performance in the target domain.

Another type of related works is semi-supervised clustering, which can be considered as a kind of knowledge-based clustering that makes use of some useful knowledge in the clustering procedure [41-45]. For example, the most commonly used knowledge is "must_link" or "should_not_link" of the data pairs [41]. In semi-supervised clustering, some prior information, such as labels of partial data, is assumed available and the data usually come from the same domain. Compared with semi-supervised clustering, the proposed TPFC algorithms are different in a sense that the data come from more than one domain.

The rest of the paper is organized as follows. In section II, several representative PFC algorithms are briefly reviewed. In section III, the concept of leveraging knowledge in clustering is introduced, and two TPFC algorithms and the corresponding extended versions are proposed. Experimental results are reported and discussed in section IV. Finally, conclusions are given in section V.

## II. PROTOTYPE-BASED CLUSTERING

Prototype-based clustering is perhaps the most extensively studied clustering technique. It typically has the following form of the objective function:

$$J = f\left(\mathbf{U}, \mathbf{\Theta}\right), \tag{1}$$

where $\mathbf{U}$ is the partition matrix and $\mathbf{\Theta}$ is the set of prototype parameters. Its aim is to optimize the objective function in Eq. (1) so as to obtain the optimal partition matrix $\mathbf{U}^*$ and the optimal prototype parameter set $\mathbf{\Theta}^*$. Let us firstly review the following two types of popular prototype-based clustering techniques, namely, cluster center prototype-based clustering [18-21] and subspace cluster prototype-based clustering [22, 51-53].

*A. Cluster Center Prototype-based Clustering*

Among the cluster center prototype-based clustering methods, two well-known fuzzy algorithms are the fuzzy c-means (FCM) clustering algorithm [19] and the possibilistic c-means (PCM) clustering algorithm [20]. Their objective functions are defined as follows.

$$\text{FCM:}\quad \min_{\mathbf{U},\mathbf{V}} J_{FCM}=\sum_{i=1}^{C}\sum_{j=1}^{N}u_{ij}^{m}\left\|\mathbf{x}_j-\mathbf{v}_i\right\|^2 \tag{2a}$$

$$st.\ u_{ij}\in[0,1],\ \sum_{i=1}^{C}u_{ij}=1,\ 0<\sum_{j=1}^{N}u_{ij}<N,$$

$$\text{PCM:}\ \min_{\mathbf{U},\mathbf{V}} J_{PCM}=\sum_{i=1}^{C}\sum_{j=1}^{N}u_{ij}^{m}\left\|\mathbf{x}_j-\mathbf{v}_i\right\|^2+\sum_{i=1}^{C}\sum_{j=1}^{N}\eta_i(1-u_{ij})^m \tag{2b}$$

$$st.\ u_{ij}\in[0,1],\ 0<\sum_{j=1}^{N}u_{ij}<N,$$

where $C$ is the number of clusters ($i=1,2,\ldots,C$); $N$ is the number of data points; $\mathbf{x}_j\in R^d$ is the jth data sample ($j=1,2,\ldots,N$); $\mathbf{V}=\left[\mathbf{v}_1,\ldots,\mathbf{v}_C\right]^T$ is the matrix of $C$ cluster centers with $\mathbf{v}_i\in R^d$; $\mathbf{U}=[u_{ij}]_{C\times N}$ is the fuzzy/possibilistic partition matrix whose element $\mu_{ij}$ denotes the membership of the *j*th data sample belonging to the *i*th class; $m$ is the fuzzy index and $\eta_i$ is a positive parameter used by PCM.

Another representative cluster center prototype-based clustering method is the maximal entropy clustering (MEC) algorithm [21] whose objective function is defined as

$$\text{MEC:}\quad \min_{\mathbf{U},\mathbf{V}} J_{MEC}=\sum_{i=1}^{C}\sum_{j=1}^{N}u_{ij}\left\|\mathbf{x}_j-\mathbf{v}_i\right\|^2+\gamma\sum_{i=1}^{C}\sum_{j=1}^{N}u_{ij}\ln(u_{ij}) \tag{2c}$$

$$st.\ u_{ij}\in[0,1],\ \sum_{i=1}^{C}u_{ij}=1,\ 0<\sum_{j=1}^{N}u_{ij}<N,$$

where $\mathbf{U}=[u_{ij}]_{C\times N}$ is the probabilistic partition matrix whose element $\mu_{ij}$ denotes the probability of the *j*th data sample belonging to the *i*th class, and $\sum_{i=1}^{C}\sum_{j=1}^{N}u_{ij}\ln(u_{ij})$ denotes the negative Shannon entropy.

*B. Subspace Cluster Prototype-based Clustering*

Subspace cluster prototype-based clustering has attracted more and more attentions in recent years [22, 51-53, 61]. It is also commonly termed as soft subspace clustering. Compared with the classical center-prototype based clustering, this type of technique has demonstrated more promising performance on high dimensional data. In this type of clustering methods, the prototype of each cluster is characterized by a cluster center and a weighting vector representing the soft subspace for this cluster. One representative algorithm is the fuzzy subspace clustering (FSC) algorithm [51] whose objective function is defined as

$$\text{FSC:}\ \min_{\mathbf{U},\mathbf{V},\mathbf{W}} J_{FSC}=\sum_{i=1}^{C}\sum_{j=1}^{N}u_{ij}\sum_{k=1}^{d}w_{ik}^{\tau}\left(x_{jk}-v_{ik}\right)^2+\sigma\sum_{i=1}^{C}\sum_{k=1}^{d}w_{ik}^{\tau} \tag{3a}$$

$$st.\ u_{ij}\in\{0,1\},\ \sum_{i=1}^{C}u_{ij}=1,\ 0<\sum_{j=1}^{N}u_{ij}<N,\ w_{ik}\in[0,1],$$

$$\sum_{k=1}^{d}w_{ik}=1,\ 0<\sum_{i=1}^{C}w_{ik}<C,$$

where $\mathbf{W}=[\mathbf{w}_1,\ldots,\mathbf{w}_C]^T$ is the matrix of weighting vectors and $\tau$ is the fuzzy index of the fuzzy weighting; $\mathbf{U}=[u_{ij}]_{C\times N}$ is the crisp partition matrix and the other parameters are defined as in Eq.(2a).

Another representative subspace cluster prototype-based clustering algorithm is the entropy weighting k-means (EWKM) algorithm [52], whose objective function is defined as

$$\text{EWKM:}\ \begin{aligned}\min_{\mathbf{U},\mathbf{V},\mathbf{W}} J_{EWKM}&=\sum_{i=1}^{C}\sum_{j=1}^{N}u_{ij}\sum_{k=1}^{d}w_{ik}\left(x_{jk}-v_{ik}\right)^2\\&+\gamma\sum_{i=1}^{C}\sum_{k=1}^{d}w_{ik}\ln(w_{ik})\end{aligned} \tag{3b}$$

$$st.\ u_{ij}\in\{0,1\},\ \sum_{i=1}^{C}u_{ij}=1,\ 0<\sum_{j=1}^{N}u_{ij}<N,\ w_{ik}\in[0,1],$$

$$\sum_{k=1}^{d}w_{ik}=1,\ 0<\sum_{i=1}^{C}w_{ik}<C,$$

where $\sum_{i=1}^{C}\sum_{k=1}^{d}w_{ik}\ln(w_{ik})$ is the negative Shannon entropy and the other parameters are defined as in (3a).

The prototype-based clustering algorithms have been studied in depth in the past decades. However, in the context of transfer learning, they have not been studied extensively as explained in Section I. In the next section, transfer learning is applied to prototype-based fuzzy clustering (PFC) and several transfer PFC (TPFC) algorithms are developed.

## III. TRANSFER PROTOTYPE-BASED FUZZY CLUSTERING

In this section, a knowledge leverage-based transfer learning mechanism is first introduced. Then two TPFC algorithms are proposed.

*A. Knowledge Leverage-based TPFC*

*1) Framework*

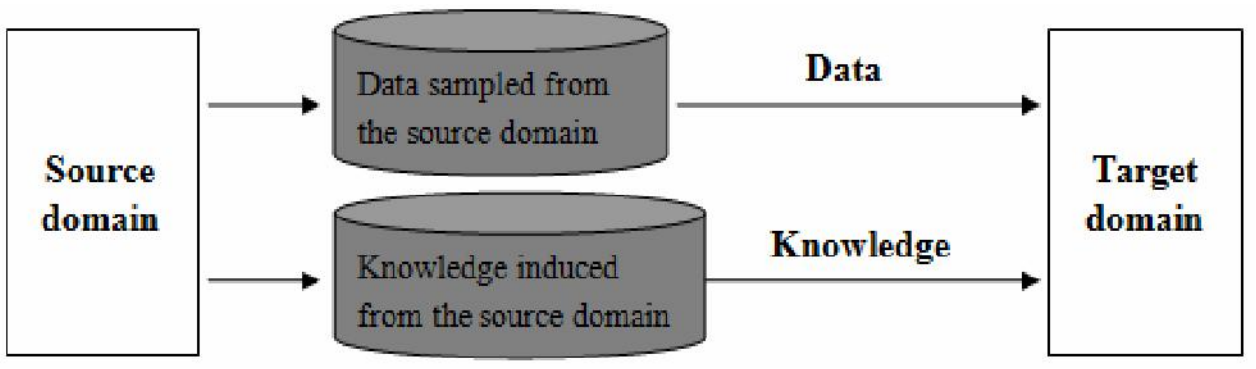

Fig.2 Useful information from the source domain for the clustering task in the target domain

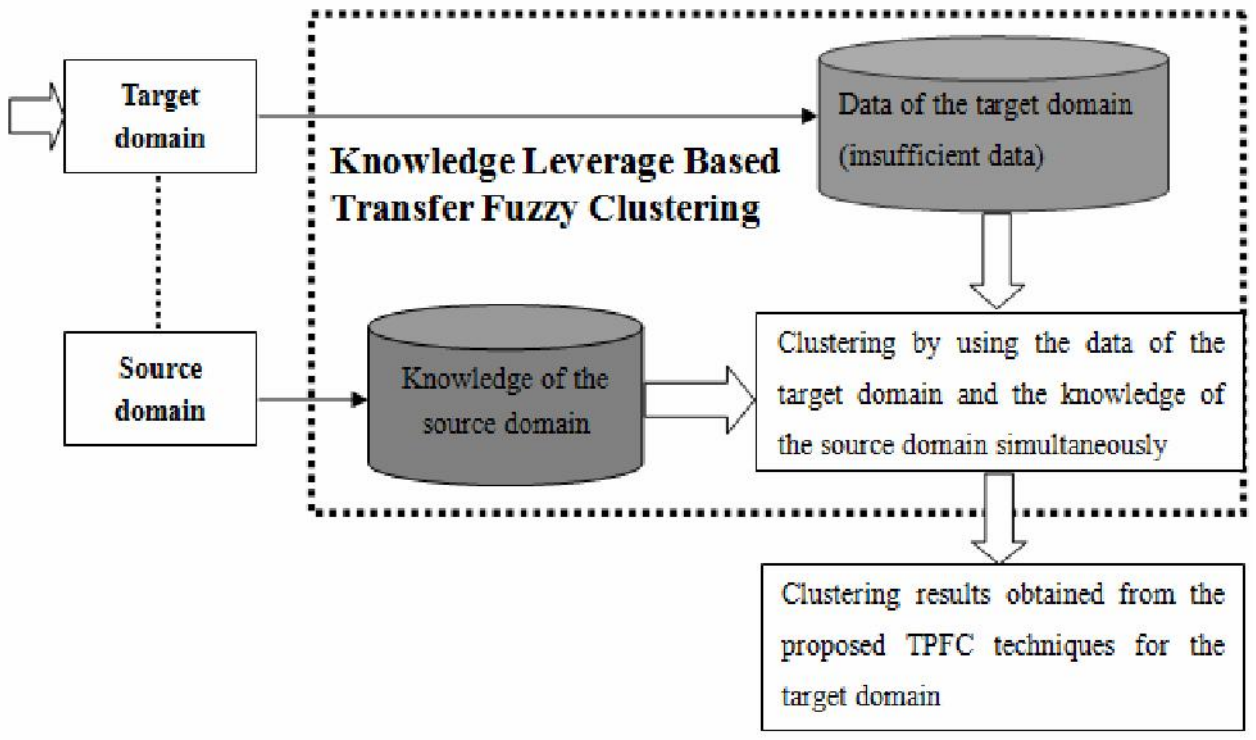

Fig. 3 Framework of the knowledge leverage-based TPFC

As mentioned in the Section I, when the data are scarce or very limited, the existing prototype-based clustering algorithms may no longer be effective. A promising strategy to cope with this challenge is to learn by reference to the source domain which might have useful information for the target domain. As shown in Fig. 2, two kinds of information are usually available from the source domain: original data and induced knowledge. It is not always appropriate to use the data from the source domain for the clustering in the target domain. In fact, it is difficult to handle the similarity and difference in data distributions between the two domains. This is because there possibly exists a drifting of data distributions between the domains. On the other hand, the influence of the induced knowledge from the source domain can be conveniently controlled because the induced knowledge is much clearer and more concise than the data in the source domain. Thus, it is more appropriate to exploit the induced knowledge rather than the data from the source domain to help the clustering task in the target domain. In this study, two TPFC algorithms using such a strategy are presented. Fig.3 depicts such a framework of knowledge leverage-based TPFC.

*2) Objective Function*

The following general objective function is proposed for TPFC based on the aforementioned knowledge leverage-based transfer learning strategy:

$$\min_{\mathbf{U},\mathbf{\Theta}} J_{TPFC} = f_t\left(\mathbf{U},\mathbf{\Theta}\right) + f_s\left(\mathbf{U},\mathbf{\Theta};\tilde{\mathbf{\Theta}}\right), \tag{4}$$

which consists of two terms. The first term $f_t\left(\mathbf{U},\mathbf{\Theta}\right)$ is inherited from the classical PFC algorithms and is used to learn from the data in the target domain. The second term $f_s\left(\mathbf{U},\mathbf{\Theta};\tilde{\mathbf{\Theta}}\right)$ is introduced to carry out knowledge leveraging from the source domain. This term can be in various forms. Its specification depends on what prototype-based clustering algorithm is formulated.

With this general objective function, two TPFC algorithms, namely, transfer fuzzy c-mean (TFCM) and transfer fuzzy subspace clustering (TFSC), are proposed in this paper firstly with an assumption that the numbers of clusters between the source domain and the target domain are the same. Furthermore, the extended versions are also presented to cater to the situation where the numbers of clusters between the source domain and the target domain are different.

In fact, transfer clustering algorithms for other prototype-based clustering algorithms such as possibilistic c-mean and entropy weighting k-mean subspace clustering can be similarly formulated. Before proceeding to present the new transfer clustering algorithms, let us explain our assumption in this paper. Our formulations of transfer clustering algorithms are based on the following assumption. The knowledge about the clusters in the source domain has been induced in advance by some learning procedures and is available for the clustering task of the target domain.

*B. TPFC Algorithms*

*1) TFCM*

Among the cluster center prototype-based clustering methods, fuzzy c-means [19] is of high popularity and is adopted here to develop our first TPFC algorithm called transfer FCM (TFCM). To do so, the following objective function is defined:

$$\begin{aligned}\min_{\mathbf{U},\mathbf{V}} J_{TFCM} &= \sum_{i=1}^{C}\sum_{j=1}^{N} u_{ij}^m \left\|\mathbf{x}_j-\mathbf{v}_i\right\|^2 + \lambda_1\cdot\sum_{i=1}^{C}\sum_{j=1}^{N} u_{ij}^m \left\|\mathbf{x}_j-\tilde{\mathbf{v}}_i\right\|^2 \\ &\quad + \lambda_2\cdot\sum_{i=1}^{C}\left(\sum_{j=1}^{N} u_{ij}^m\right)\left\|\tilde{\mathbf{v}}_i-\mathbf{v}_i\right\|^2 \end{aligned} \tag{5}$$

$$st.\ u_{ij}\in[0,1],\ \sum_{i=1}^{C} u_{ij}=1,\ 0<\sum_{j=1}^{N} u_{ij}<N,$$

where $\tilde{\mathbf{v}}_i$ denotes the $i$th cluster center obtained in the source domain, and $\lambda_1$ and $\lambda_2$ are non-negative parameters.

The concepts behind this knowledge leverage-based transfer learning mechanism can be given as follows.

(1) The first term in Eq. (5) is directly inherited from the classical FCM, which is mainly used to learn from the data available in the target domain.

(2) The second and third terms are used to learn with the knowledge from the source domain, where the knowledge is the cluster centers obtained in the source domain for a certain clustering task.

(3) By having the second term, the clustering procedure is apt to obtain a partition matrix by using the cluster centers obtained in the source domain, which implies the use of a supervised learning procedure, such as the one-nearest-neighbor (1-NN) classification, in the target domain.

(4) By having the third term, the desired cluster centers will be apt to approximate the ones obtained in the source domain, and thus guiding the clustering procedure to learn from the knowledge of the source domain.

(5) The parameters $\lambda_1$ and $\lambda_2$, are used to balance the influence of data in the target domain and knowledge in the source domain. Parameter analysis will be given in our experimental section.

By using a similar optimization strategy in FCM, one can obtain the following learning rules based on Eq. (5).

$$\mathbf{v}_i = \frac{\sum_{j=1}^{N} u_{ij}^m \mathbf{x}_j + \lambda_2 \sum_{j=1}^{N} u_{ij}^m \tilde{\mathbf{v}}_i}{\sum_{j=1}^{N} u_{ij}^m + \lambda_2 \sum_{j=1}^{N} u_{ij}^m}, \tag{6}$$

$$u_{ij} = \frac{\left(\dfrac{1}{\left\|\mathbf{x}_j-\mathbf{v}_i\right\|^2+\lambda_1\left\|\mathbf{x}_j-\tilde{\mathbf{v}}_i\right\|^2+\lambda_2\left\|\tilde{\mathbf{v}}_i-\mathbf{v}_i\right\|^2}\right)^{1/(m-1)}}{\sum\limits_{k=1}^{C}\left[\dfrac{1}{\left\|\mathbf{x}_j-\mathbf{v}_k\right\|^2+\lambda_1\left\|\mathbf{x}_j-\tilde{\mathbf{v}}_k\right\|^2+\lambda_2\left\|\tilde{\mathbf{v}}_k-\mathbf{v}_k\right\|^2}\right]^{1/(m-1)}}. \tag{7}$$

Based on Eqs.(6) and (7), the TFCM algorithm is described below.

***TFCM Algorithm***

1) Initialize the number of iterations as $t=0$ and the partition matrix $\mathbf{U}(0)$ randomly; Set the maximum number of iterations $t_{\max}$ and threshold $\varepsilon$; Set the balance parameters $\lambda_1$, $\lambda_2$.

2) Update the matrix of the cluster centers $\mathbf{V}(t)$ using (6);

3) Set $t=t+1$;

4) Update the partition matrix $\mathbf{U}(t)$ using (7);

5) If $\left\|\mathbf{U}(t)-\mathbf{U}(t\text{-}1)\right\|<\varepsilon$ or $t=t_{\max}$, then terminate; else go to 2).

For other popular center prototype-based clustering algorithms, such as PCM [20], one can develop the corresponding transfer learning versions in a similar manner.

*2) TFSC*

Among the existing subspace cluster prototype-based clustering methods, the FSC algorithm [51] is adopted to develop the corresponding transfer learning version, i.e., TFSC. Thus, the following objective function is proposed:

$$\begin{aligned}\min_{\mathbf{U},\mathbf{V},\mathbf{W}} J_{TFSC} = & \sum_{i=1}^{C}\sum_{j=1}^{N}u_{ij}^{m}\sum_{k=1}^{d}w_{ik}^{\tau}\left(x_{jk}-v_{ik}\right)^2+\sigma\cdot\sum_{i=1}^{C}\sum_{k=1}^{d}w_{ik}^{\tau} \\ & +\lambda_1\cdot\sum_{i=1}^{C}\sum_{j=1}^{N}u_{ij}^{m}\sum_{k=1}^{d}\tilde{w}_{ik}^{\tau}\left(x_{jk}-\tilde{v}_{ik}\right)^2+\lambda_2\cdot\sum_{i=1}^{C}(\sum_{j=1}^{N}u_{ij}^{m})\sum_{k=1}^{d}\tilde{w}_{ik}^{\tau}\left(\tilde{v}_{ik}-v_{ik}\right)^2\end{aligned} \quad (8)$$

$st.$ $u_{ij}\in[0,1]$, $\sum_{i=1}^{C}u_{ij}=1$, $0<\sum_{j=1}^{N}u_{ij}<N$, $w_{ik}\in[0,1]$, $\sum_{k=1}^{d}w_k=1$.

In Eq.(8), the first two terms are directly inherited from the classical FSC algorithm except that we have extended the hard partition $u_{ij}\in\{0,1\}$ in FSC to the fuzzy partition $u_{ij}\in[0,1]$ for easy optimization and knowledge leverage. The third and fourth terms are used to learn from the knowledge of the source domain. The third term implies that we can learn the partition matrix by referring to the subspace cluster prototypes obtained in the source domain, characterized by the cluster centers and weighting vectors, i.e., $\tilde{v}_{ik}$ and $\tilde{w}_{ik}^{\tau}$. The fourth term intends to make the desired cluster centers apt to approximate the ones obtained in the source domain under the subspace constraints for the source domain. Thus, by minimizing Eq. (8), the clustering procedure can learn from the data of the target domain and the knowledge of the source domain simultaneously. Again, by following the optimization strategy in FSC, we can obtain the following learning rules:

$$v_{ik}=\frac{\sum_{j=1}^{N}u_{ij}^{m}w_{ik}^{\tau}x_{jk}+\lambda_2\cdot\sum_{j=1}^{N}u_{ij}^{m}\tilde{w}_{ik}^{\tau}\tilde{v}_{ik}}{\sum_{j=1}^{N}u_{ij}^{m}w_{ik}^{\tau}+\lambda_2\cdot\sum_{j=1}^{N}u_{ij}^{m}\tilde{w}_{ik}^{\tau}}, \quad (9)$$

$$w_{ik}=\left[\frac{1}{\sum_{j=1}^{N}u_{ij}^{m}\left(x_{jk}-v_{ik}\right)^2+\sigma}\right]^{1/(\tau-1)}\Bigg/\sum_{s=1}^{d}\left[\frac{1}{\sum_{j=1}^{N}u_{ij}^{m}\left(x_{js}-v_{is}\right)^2+\sigma}\right]^{1/(\tau-1)}, \quad (10)$$

$$u_{ij}=\left[1/d_{ij}\right]^{1/(m-1)}\Big/\sum_{k=1}^{C}\left[\left(1/d_{ik}\right)\right]^{1/(m-1)}, \quad (11)$$

$$\begin{aligned}d_{ik}= & \sum_{k=1}^{d}w_{ik}^{\tau}\left(x_{jk}-v_{ik}\right)^2+\lambda_1\cdot\sum_{k=1}^{d}\tilde{w}_{ik}^{\tau}\left(x_{jk}-\tilde{v}_{ik}\right)^2 \\ & +\lambda_2\cdot\sum_{k=1}^{d}\tilde{w}_{ik}^{\tau}\left(\tilde{v}_{ik}-v_{ik}\right)^2\end{aligned}. \quad (12)$$

Based on Eqs.(9)-(12), the TFSC algorithm can be described below.

***TFSC Algorithm***

1) Initialize the number of iterations as $t=0$ and the partition matrix $\mathbf{U}(0)$ randomly; Set the maximum number of iterations $t_{max}$ and threshold $\varepsilon$; Set the balance parameters $\lambda_1$, $\lambda_2$.
2) Update the matrix of the cluster centers $\mathbf{V}(t)$ using (9);
3) Update the matrix of the weighting vectors $\mathbf{W}(t)$ using (10);
4) Set $t=t+1$;
5) Update the partition matrix $\mathbf{U}(t)$ using (11);
6) If $\|\mathbf{U}(t)-\mathbf{U}(t\text{-}1)\|<\varepsilon$ or $t$=$t_{max}$, then terminate; else go to 2).

***Remark***: According to the third term of Eq. (5), and the fourth term of Eq. (8), not only the number of clusters should be the same between the source domain and the target domain, but also each cluster should be perfectly matched. When imperfect matching happens, the performance of the whole clustering procedure will perhaps deteriorate a lot. This issue may occur in some situations, for example, when a bad partition matrix is randomly initialized or the number of clusters is too many. The imperfect matching of the cluster centers between the two domains has a negative influence on the performance of transfer clustering. In particular, if less knowledge from the source domain can be used, the negative influence will become larger. However, if more knowledge from the source domain can be adopted, the negative influence will become weaker accordingly. In order to explain this issue, let us take TFCM as a case to study. In TFCM, from Eq.(6), we can see that when less knowledge from the source domain is used, i.e., smaller value is set for the parameter $\lambda_2$, $\mathbf{v}_i$'s are apt to approximate the values obtained by only the data in the target domain. For example, if $\lambda_2\to 0$, $\mathbf{v}_i\to\sum_{j=1}^{N}u_{ij}^{m}\mathbf{x}_j\Big/\sum_{j=1}^{N}u_{ij}^{m}$. Thus, the bad initialization of the partition matrix $\mathbf{U}=[u_{ij}]$ may result in having imperfect matching of cluster centers between the two domains and consequently the clustering results will be deteriorated. However, we find that when more knowledge from the source domain is used, i.e., larger values are set for the parameters $\lambda_1$ and $\lambda_2$ in TFCM, the occurrence of imperfect matching can be avoided to a large extent due to the following fact. From Eq.(6), we can see that $\mathbf{v}_i$ will be apt to approximate the values of the cluster centers obtained in the source domain when the value of the parameter $\lambda_2$ is large. For example, if $\lambda_2\to\infty$, then $\mathbf{v}_i\to\tilde{\mathbf{v}}_i$. Meanwhile, from Eq.(7) we can see that when the value of parameter $\lambda_1$ is large, $u_{ij}\approx\left(\frac{1}{\|\mathbf{x}_j-\tilde{\mathbf{v}}_i\|^2}\right)^{1/(m-1)}\Bigg/\left[\sum_{k=1}^{C}\frac{1}{\|\mathbf{x}_j-\tilde{\mathbf{v}}_k\|^2}\right]^{1/(m-1)}$, which implies that if a data point is near to the cluster center $\tilde{\mathbf{v}}_i$ in the source domain, it is apt to be classified as the matching cluster in the target domain with the cluster center $\mathbf{v}_i$. This fact indicates that more knowledge from the source domain will really reduce the chance of imperfect matching of the cluster centers between the two domains for the proposed algorithms. According to the above analysis, one may conclude that a good solution to avoid imperfect matching is very useful. A feasible method is to give a good initialization for the partition matrix in the target domain by using the knowledge and the rule obtained in the source domain. For example, for TFCM, the initial partition matrix can be computed by using the data in the target domain and the cluster centers obtained in the source domain based on the update rule of the membership function in FCM. In particular, an example of clustering a dataset consisting of 20 clusters is presented in subsection IV-E to demonstrate the effectiveness of such a strategy.

*C. Extended TPFC Algorithms*

Although the proposed two TPFC algorithms, i.e., TFCM

and TFSC, can effectively implement transfer learning for the clustering in the target domain, they both have a common limitation, i.e., the numbers of clusters in the source and target domains must be the same. However, in most practical applications the above assumption cannot always hold. Thus, the development of TPFC algorithm that can implement the transfer clustering with different numbers of clusters in the source and target domains are very necessary. In this subsection, two extended TPFC algorithms, i.e., extended transfer FCM (E-TFCM) and extended transfer FSC (E-TFSC), are proposed for this purpose.

*1) E-TFCM*

Based on the objective function of TFCM in (5), we find it is difficult to directly extend it for the scene with different numbers of clusters in both domains. Instead, the following objective function is used for E-TFCM

$$\min_{\mathbf{U,V,R}} J_{\text{E-TFCM}} = \sum_{i=1}^{C_t}\sum_{j=1}^{N} u_{ij}^{m_1}\|\mathbf{x}_j-\mathbf{v}_i\|^2 + \lambda_1\cdot\sum_{i=1}^{C_t}\sum_{l=1}^{C_s} r_{il}^{m_2}\|\tilde{\mathbf{v}}_l-\mathbf{v}_i\|^2 \quad (13)$$

$$st.\ u_{ij}\in[0,1],\ \sum_{i=1}^{C_t}u_{ij}=1,\ 0<\sum_{j=1}^{N}u_{ij}<N,\ r_{il}\in[0,1],\ \sum_{l=1}^{C_s}r_{il}=1,\ 0<\sum_{l=1}^{C_s}r_{il}<C_s$$

For Eq. (13), the following explanations are given.

(1) The first term in (13) is directly inherited from the classical FCM, which is mainly used to learn from the data available in the target domain.

(2) The second one is used to learn with the knowledge from the source domain. In this term, $r_{il}$ denotes the similarity between the $ith$ cluster in the target domain and the $lth$ cluster in the source domain; this term implies that if the $ith$ cluster in the target domain and the $lth$ cluster in the source domain are more similar, the $ith$ cluster in the target domain will learn more knowledge from the $lth$ cluster in the source domain.

By following the optimization strategy in TFCM, we can obtain the following update rules for E-TFCM.

$$\mathbf{v}_i = \frac{\sum_{j=1}^{N}u_{ij}^{m_1}\mathbf{x}_j + \lambda_1\sum_{l=1}^{C_s}r_{il}^{m_2}\tilde{\mathbf{v}}_l}{\sum_{j=1}^{N}u_{ij}^{m_1} + \lambda_1\sum_{l=1}^{C_s}r_{il}^{m_2}}, \quad (14)$$

$$u_{ij} = \left(\frac{1}{\|\mathbf{x}_j-\mathbf{v}_i\|^2}\right)^{1/(m_1-1)} \Bigg/ \sum_{k=1}^{C_t}\left[\frac{1}{\|\mathbf{x}_j-\mathbf{v}_k\|^2}\right]^{1/(m_1-1)}, \quad (15)$$

$$r_{il} = \frac{1}{\sum_{q=1}^{C_s}\left\|\tilde{\mathbf{v}}_l-\mathbf{v}_i\right\|^2 \Big/ \left\|\tilde{\mathbf{v}}_q-\mathbf{v}_i\right\|^{2\frac{1}{m_2-1}}}. \quad (16)$$

Based on Eqs.(14)-(16), the E-TFCM algorithm can be easily obtained. Here the details of the E-TFCM algorithms are not given for saving the space of the paper.

*2) E-TFSC*

Based on the objective function of TFSC, we find it is difficult to directly extend it for the scene with different numbers of clusters in both domains. However, we can adopt the following objective function to develop the E-TFSC.

$$\min_{\mathbf{U,V,W}} J_{TFSC} = \sum_{i=1}^{C_t}\sum_{j=1}^{N}u_{ij}^{m_1}\sum_{k=1}^{d}w_{ik}^{\tau}\left(x_{jk}-v_{ik}\right)^2 + \sigma\cdot\sum_{i=1}^{C_t}\sum_{k=1}^{d}w_{ik}^{\tau} + \lambda_1\cdot\sum_{i=1}^{C_t}\sum_{l=1}^{C_s}r_{il}^{m_2}\sum_{k=1}^{d}\tilde{w}_{lk}^{\tau}\left(\tilde{v}_{lk}-v_{ik}\right)^2 \quad (17)$$

$$st.\ u_{ij}\in[0,1],\ \sum_{i=1}^{C_t}u_{ij}=1,\ 0<\sum_{j=1}^{N}u_{ij}<N,\ r_{il}\in[0,1],\ 0<\sum_{l=1}^{C_s}r_{il}<C_s\ \ w_{ik}\in[0,1],\ \sum_{k=1}^{d}w_k=1.$$

For Eq. (17), some explanations are also given below as for Eq.(13). The first two terms in (17) are directly inherited from the classical FSC algorithm except that we have extended the hard partition $u_{ij}\in\{0,1\}$ in FSC to the fuzzy partition $u_{ij}\in[0,1]$ for easy optimization and knowledge leverage. The first term is mainly used to learn from the data available in the target domain. The third term is used to learn with the knowledge from the source domain, where $r_{il}$ denotes the similarity between the $ith$ cluster in the target domain and the $lth$ cluster in the source domain. With the third term, if the $ith$ cluster in the target domain and the $lth$ cluster in the source domain are more similar, it will learn more knowledge from the $lth$ cluster in the source domain.

By following the optimization strategy in TFSC, we can obtain the following learning rules for E-TFSC.

$$v_{ik} = \frac{\sum_{j=1}^{N}u_{ij}^{m_1}w_{ik}^{\tau}x_{jk} + \lambda_1\cdot\sum_{l=1}^{C_s}r_{il}\tilde{w}_{lk}^{\tau}\tilde{v}_{lk}}{\sum_{j=1}^{N}u_{ij}^{m_1}w_{ik}^{\tau} + \lambda_1\cdot\sum_{l=1}^{C_s}r_{il}\tilde{w}_{lk}^{\tau}}, \quad (18)$$

$$w_{ik} = \frac{\left[\dfrac{1}{\sum_{j=1}^{N}u_{ij}^{m_1}\left(x_{jk}-v_{ik}\right)^2+\sigma}\right]^{1/(\tau-1)}}{\sum_{s=1}^{d}\left[\dfrac{1}{\sum_{j=1}^{N}u_{ij}^{m_1}\left(x_{js}-v_{is}\right)^2+\sigma}\right]^{1/(\tau-1)}}, \quad (19)$$

$$u_{ij} = \left[1/d_{ij}\right]^{1/(m_1-1)} \Big/ \sum_{k=1}^{C_t}\left[(1/d_{ik})\right]^{1/(m_1-1)}, \quad (20)$$

$$d_{ik} = \sum_{k=1}^{d}w_{ik}^{\tau}\left(x_{jk}-v_{ik}\right)^2, \quad (21)$$

$$r_{il} = \frac{1}{\sum_{q=1}^{C_s}\left(\dfrac{\sum_{k=1}^{d}\tilde{w}_{lk}^{\tau}\left(\tilde{v}_{lk}-v_{ik}\right)^2}{\sum_{k=1}^{d}\tilde{w}_{qk}^{\tau}\left(\tilde{v}_{qk}-v_{ik}\right)^2}\right)^{\frac{1}{m_2-1}}}. \quad (22)$$

Based on Eqs.(18)-(22), the E-TFCM algorithm can be easily obtained. Here the details are not given for saving the space of the paper.

***Remark***: Here some relations between TFCM and E-TFCM are discussed. While TFCM is only suitable for the scene that the number of clusters between the source domain and target domain must be the same, E-TFCM can be used when the numbers of clusters in both domains are different. However, it is noted here that TFCM is not a special case of E-TFCM. Although both TFCM and E-TFCM are applicable to the situation that the numbers of clusters in both domain are the same, the former is more advantageous to the latter. It is due to the fact that the former has used two knowledge transfer items for transfer learning but the latter only adopt one term.

Thus, when the numbers of cluster in both domains are the same, TFCM will have stronger transfer learning abilities than E-TFCM. For TFSC and E-TFSC, their relation is similar to that between TFCM and E-TFCM.

*D. Computational Complexity and Convergence Analysis*

The computational complexities of the proposed algorithms can be described as follows. For the TFCM and E-TFCM algorithms, their computational complexities are $O(TNC+TC)$, where $T$ is the total number of iterations; $N$ is the size of a dataset (i.e., the number of samples), and $C$ is the number of clusters. In other words, they are of the same order as those of the classical FCM algorithm. For the TFSC and E-TFSC algorithms, their computational complexities are $O(TNC+TC+TCd)$, where $d$ is the number of features, which is of the same order as that of the classical FSC.

For the convergence of the proposed four TPFC algorithms. The convergence can be proved based on Zangwill's convergence theorem [54, 55] in a similar manner in [51]. Note here that similar to the FCM-like methods, only a local optimal solution can be obtained since the optimization problems of the concerned four algorithms above are non-convex. However, this usually does not lead to a serious problem as the local optimal solution is effective enough in most practical applications.

## IV. Experimental Results

In this section, the proposed algorithms will be extensively evaluated on synthetic and real-world datasets. The indices used for performance evaluation and the experimental setup are first described. Then, the performance of the proposed algorithms on synthetic and real-world text datasets is reported and discussed. Further comprehensive comparison with other related algorithms is conducted on several synthetic and real-world datasets. All the algorithms were implemented with Matlab codes and experiments were run on a computer with 1.66GHz CPU and 2GB RAM.

*A. Performance Indices and Experimental Setup*

Two metrics, i.e., the *rand index* (RI) and the *normalized mutual information* (NMI) [56], are used to evaluate the performance of the clustering algorithms. RI is commonly defined as

$$RI=\frac{f_{00}+f_{11}}{N(N-1)/2}$$

where $f_{00}$ is the number of pairs of data points having different class labels and belonging to different clusters; $f_{11}$ is the number of pairs of data points having the same class labels and belonging to the same clusters; $N$ is the size of the whole dataset. NMI is defined and computed according to the following formula:

$$NMI=\frac{\sum_{i=1}^{C}\sum_{j=1}^{C}N_{i,j}\log\frac{N\cdot N_{i,j}}{N_i\cdot N_j}}{\sqrt{\sum_{i=1}^{C}N_i\log\frac{N_i}{N}\cdot\sum_{j=1}^{C}N_j\log\frac{N_j}{N}}},$$

where $N_{i,j}$ is the number of agreements between cluster $i$ and class $j$; $N_i$ is the number of data points in cluster $i$; $N_j$ is the number of data points in class $j$, and $N$ is the size of the whole dataset. Both RI and NMI take a value within the interval [0, 1]. The higher the values, the better the clustering performance is.

When RI and NMI are computed, the prior labels of the data must be available. In addition to them, an index that does not require the labels of data is also used for performance evaluation. Since most of the adopted methods for comparison are fuzzy measure-based methods, the fuzzy validity measure is naturally very appropriate. Here, the classical fuzzy validity measure, i.e., Xie-Beni (XB) index, is adopted to further evaluate and compare the performance of different fuzzy measure-based algorithms. The adopted XB index for FCM is defined as follows:

$$XB(\mathbf{U},\mathbf{V};\mathbf{X})=\frac{\frac{1}{N}\sum_{i=1}^{C}\sum_{j=1}^{N}u_{ij}^{m}\left\|\mathbf{x}_j-\mathbf{v}_i\right\|^2}{\min_{i\neq j}\left\{\|\mathbf{v}_i-\mathbf{v}_j\|\right\}}.$$

Usually, the smaller the index values, the better the clustering performance is.

As the *XB* index was originally designed for FCM-like algorithms only, another modified *XB* index was presented to evaluate the subspace cluster prototype-based algorithms here. For FSC, TFSC and E-TFSC, another modified *XB* index which includes the subspace information is proposed:

$$XB_{FSC}(\mathbf{U},\mathbf{V},\mathbf{W};\mathbf{X})=\frac{\frac{1}{N}\sum_{i=1}^{C}\sum_{j=1}^{N}u_{ij}^{m}\sum_{k=1}^{d}w_{ik}^{\tau}\left(x_{jk}-v_{ik}\right)^2}{\min_{i\neq j}\left\{\sum_{k=1}^{d}w_{ik}^{\tau}\left(v_{ik}-v_{jk}\right)^2\right\}}.$$

Similar to the *XB*, the smaller the index values of the $XB_{FSC}$, the better the clustering performance is.

The proposed transfer clustering algorithms are compared with the corresponding non-transfer learning counterparts in subsections IV-B & IV-C. In sub-section IV-D, a comparison with several related learning algorithms is presented. In subsection IV-E, some discussions are presented. In our experiments, the clustering process of all algorithms is repeated 20 times for each parameter setting on a dataset. The maximum number of iterations, i.e., $t_{\max}$, is set to 100 for all iteration-based clustering algorithms. For the proposed algorithms, the involved fuzzy indices, i.e., $m$, $m_1$ and $m_2$, are set with the following rule proposed in [62]: If $d\le 2$, $m=m_1=m_2=2$; Otherwise, $m=m_1=m_2=\frac{\min(N,d-1)}{(\min(N,d-1)-2)}$.

Here, $N$ and $d$ are the number of samples and the number of dimensionality, respectively.

*B. Synthetic Datasets*

In this study, several synthetic datasets are generated to evaluate the performance of the proposed algorithms. In particular, we focus on the following aspects in this subsection: (1) how to determine the appropriate parameters for the proposed algorithms; (2) how to control the negative transfer of the bad knowledge in the source domain; (3) performance evaluation on datasets where the numbers of clusters in the source and target domains are different; and (4) performance evaluation on the high dimensional datasets.

*1) How to Determine the Appropriate Parameters*

The parameter settings are very important for the proposed four TPFC algorithms. In this subsection, we will take TFCM as the example to investigate the strategy of parameter selection. For this purpose and the study of negative transferring problem, three synthetic datasets, denoted as S1-1, S1-2 and T1, with predetermined cluster structures are used for parameter analysis of the TFCM algorithm. T1 is generated as the dataset in the target domain. S1-1 and S1-2 datasets are generated as the datasets in a good source domain and a bad source domain, respectively. Each dataset contains three clusters with different Gaussian distributions as shown in Fig 4. The parameters used to generate the datasets are listed in Table I. From Table I, we can see that three clusters in S1-1 and T1 have similar but different means and covariances, which means that S1-1 is a source domain dataset containing useful information for clustering dataset T1 in the target domain. Meanwhile, we can see that three clusters in S1-2 and T1 have very different means and covariances, which indicates that S1-2 is a source domain dataset containing bad information for clustering the target domain dataset T1.

TABLE I
PARAMETERS USED TO GENERATE THE SYNTHETIC DATASETS FOR PARAMETER ANALYSES OF THE TFCM ALGORITHM

| | Two Source domains | | | | | |
|---|---|---|---|---|---|---|
| | Good source domain (S1-1) | | | Bad source domain (S1-2) | | |
| Cluster | Cluster-1 | Cluster-2 | Cluster-3 | Cluster-1 | Cluster-2 | Cluster-3 |
| **u** | [−5 , 6] | [0 , 15] | [5, 5] | [-2 , 10] | [6, 7] | [−1 , 3] |
| **Σ** | $\begin{bmatrix}5,0\\0,5\end{bmatrix}$ | $\begin{bmatrix}5,0\\0,5\end{bmatrix}$ | $\begin{bmatrix}5,0\\0,5\end{bmatrix}$ | $\begin{bmatrix}3,0\\0,3\end{bmatrix}$ | $\begin{bmatrix}3,0\\0,3\end{bmatrix}$ | $\begin{bmatrix}3,0\\0,3\end{bmatrix}$ |
| **Size** | 600 | | | 600 | | |
| | Target domain (T1) | | | | | |
| Cluster | Cluster-1 | Cluster-2 | Cluster-3 | | | |
| **u** | [−5 , 3] | [0 , 13] | [4, 3] | | | |
| **Σ** | $\begin{bmatrix}8,0\\0,8\end{bmatrix}$ | $\begin{bmatrix}8,0\\0,8\end{bmatrix}$ | $\begin{bmatrix}8,0\\0,8\end{bmatrix}$ | | | |
| **Size** | 80 | | | | | |

* **u** and **Σ** denote the mean and covariance of each cluster, respectively.

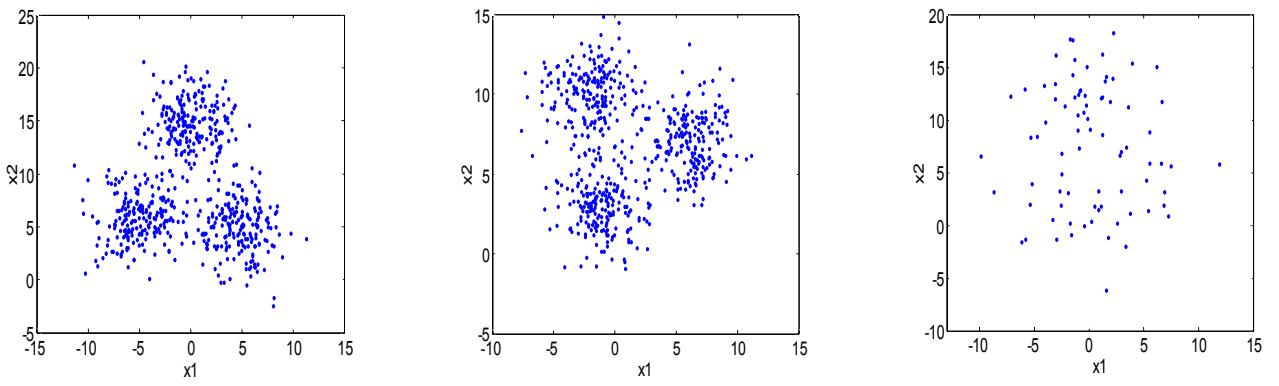

(a) S1-1 (b) S1-2 (c) T1

Fig.4 Synthetic datasets used for parameter analysis of the TFCM algorithm: (a) dataset S1-1 consisting of three clusters in a good source domain; (b) dataset S1-2 consisting of three clusters in a bad source domain; (c) dataset T1 consisting of three clusters in the target domain.

Tables II-IV record the means and standard deviations of the *RI*, *NMI* and *XB* values on the target dataset T1 obtained by the proposed TFCM algorithm using different parameter settings based on the knowledge extracted from the dataset S1-1 in a good source domain. Based on these results, the following observations can be obtained.

(1) When the two parameters in Eq.(5) are set as $\lambda_1 = 0$ and $\lambda_2 = 0$, the TFCM algorithm obtains the worst clustering effect as indicated by the mean *NMI*, *RI* and *XB* values in Part A of Tables II and IV, while the clustering performance has been improved when $\lambda_1$ and $\lambda_2$ are set to different positive values. Here, when $\lambda_1 = 0$ and $\lambda_2 = 0$ , TFCM is degenerated to the classical FCM algorithm. Thus the results in Part A of Tables II, III and IV confirm the advantage of knowledge leverage in our transfer clustering algorithm.

(2) From Part B of Tables II, III and IV, we can see that when the knowledge of the source domain is used, the standard deviations of the three indices (*RI*, *NMI* and *XB*) have been reduced. This implies that the introduction of the knowledge of the source domain has enhanced the stability/robustness of the algorithm.

(3) Tables II-IV show that the transfer of knowledge of the source domain to the clustering in the target domain has to be properly controlled. Although the knowledge of the source domain can improve the clustering performance, the improvement will be degraded if very large $\lambda_1$ and $\lambda_2$ values are used. From our experimental results, setting $\lambda_1 \in [0.1, 0.7]$ and $\lambda_2 \in [0.1, 0.7]$ can obtain more promising clustering performance on the adopted synthetic dataset T1.

From the above analysis, we know that for the proposed TFCM algorithm, the parameter settings are very important in order to obtain the better performance. Thus, a proper evaluation strategy is necessary to determine the appropriate parameters. For this purpose, the following analysis and conclusions are given.

(1) Unlike the supervised learning, no supervised information can be used by parameter selection strategies, such as the commonly used cross validation strategy for the supervised learning algorithms. For most unsupervised learning algorithms, it is still an open problem to determine the optimal values for the parameters involved. In fact, the optimal parameters are usually dependent on applications. Thus, for a specific application, users can determine an appropriate range for the involved parameters of the adopted algorithm by the prior knowledge or the available validation dataset.

(2) Tables II-IV show that the trends of the clustering results with the three indices are similar under different parameter settings, i.e., when the value of one index is better, usually the other two are also better. In fact, we have made many different experiments on different datasets with the proposed four algorithms, the similar observations have been obtained. Thus, based on this observation, a promising strategy to determine the appropriate values for the parameters involved is proposed as follows. *For the proposed TFCM and E-TFCM, the XB index can be adopted as a evaluation index to determine the appropriate parameters, while for the proposed TFSC and E-TFSC, the improved* $XB_{FSC}$ *index is a feasible index to determine the appropriate parameters*. The above conclusion can be explained based on the following facts: (i) the *XB* and modified *XB* indices, unlike the RI and NMI indices, can be computed without the supervised label information; (ii) in most situations when the XB index or the modified XB index is better in a certain parameter setting, the obtained *RI* and *NMI* values are usually much better accordingly.

(3) In addition, in order to weaken the influence of the parameters, some other investigations can also be done as suggested in [59, 60]. For example, the ensemble learning technique seems to be a promising one. By integrating the clustering results obtained under different parameter settings, ensemble clustering will provide comparable results with the case under the optimal parameter setting [59, 60].

Based on the above analysis, in the subsequent experiments, we have used the XB index to determine approximate parameters for TFCM and E-TFCM, and used the $XB_{FSC}$ index to determine approximate parameters for TFSC and E-TFSC, respectively.

***Remark***: An interesting result is that in Tables II-IV, the scores of the table are nearly symmetrical. Some explanations are given here. In (5), two knowledge transfer items have been introduced for TFCM. Since both items are constructed based on the same knowledge of the source domain, i.e., the cluster centers obtained in the source domain, they have shown the similar functions for knowledge leverage. For example, when $\lambda_1 = 0$ and $\lambda_2$ is high, the results are the same to the results obtained when $\lambda_2 = 0$ and $\lambda_1$ is high. However, the physical meanings of these two items are different, which makes their contributions to be not the same to a certain extent. Thus, we expect that the optimal knowledge transfer can be obtained with the best combination of these two parameters,. For example, in Tables II-IV, with $\lambda_1$=0.1 and $\lambda_2$=0.5 the best results are obtained. In fact, the knowledge transfer items can be developed with various ways and this is an open problem. On one hand, we hope more knowledge can be introduced in the objective function; on the other hand, we must consider the involved optimization problem in order to easily solve the objective function. Although some feasible ways have been tried in this study, many works can be addressed in depth.

In Table V, the optimized results of TFCM on the dataset T1 with the dataset S1-1 as the source domain dataset are reported, where the values of parameters were determined with the XB index as the evaluation criterion. It shows that the optimized clustering results of TFCM are obviously superior to those of the classical FCM algorithm on dataset T1. For example, in terms of the NMI index, TFCM obtains the mean value 0.7850 and the standard deviation 1.17e-16 for the data T1, while the corresponding results by FCM are 0.6561 and 0.0536.

TABLE II

MEANS AND STANDARD DEVIATIONS OF NMI OBTAINED BY TFCM ON DATASET T1 IN TARGET DOMAIN BASED ON KNOWLEDGE OF DATASET S1-1 IN THE SOURCE DOMAIN UNDER DIFFERENT PARAMETER SETTINGS

| $\lambda_2$ \ $\lambda_1$ | 0 | 0.005 | 0.1 | 0.5 | 0.7 | 1 | 1.5 | 10 | 50 | 100 |
|---|---|---|---|---|---|---|---|---|---|---|
| Part A: Means | | | | | | | | | | |
| 0 | 0.6561 | 0.6584 | 0.6932 | 0.7754 | 0.7200 | 0.6929 | 0.6929 | 0.6929 | 0.6929 | 0.6929 |
| 0.005 | 0.6663 | 0.6621 | 0.6932 | 0.7800 | 0.7200 | 0.6929 | 0.6929 | 0.6929 | 0.6929 | 0.6929 |
| 0.1 | 0.6735 | 0.6932 | 0.7801 | 0.7807 | 0.7251 | 0.6929 | 0.6929 | 0.6929 | 0.6929 | 0.6929 |
| 0.5 | 0.7760 | 0.7835 | **0.7850** | 0.7251 | 0.7251 | 0.6929 | 0.6929 | 0.6929 | 0.6929 | 0.6929 |
| 0.7 | 0.7200 | 0.7200 | 0.7251 | 0.7251 | 0.7251 | 0.6929 | 0.6929 | 0.6929 | 0.6929 | 0.6929 |
| 1 | 0.6929 | 0.6929 | 0.6929 | 0.6929 | 0.6929 | 0.6929 | 0.6929 | 0.6929 | 0.6929 | 0.6929 |
| 1.5 | 0.6929 | 0.6929 | 0.6929 | 0.6929 | 0.6929 | 0.6929 | 0.6929 | 0.6929 | 0.6929 | 0.6929 |
| 10 | 0.6929 | 0.6929 | 0.6929 | 0.6929 | 0.6929 | 0.6929 | 0.6929 | 0.6929 | 0.6929 | 0.6929 |
| 50 | 0.6929 | 0.6929 | 0.6929 | 0.6929 | 0.6929 | 0.6929 | 0.6929 | 0.6929 | 0.6929 | 0.6929 |
| 100 | 0.6929 | 0.6929 | 0.6929 | 0.6929 | 0.6929 | 0.6929 | 0.6929 | 0.6929 | 0.6929 | 0.6929 |
| Part B: Standard deviations | | | | | | | | | | |
| 0 | 0.0536 | 0.0217 | 1.17e-16 | 1.17e-16 | 1.17e-16 | 1.17e-16 | 1.17e-16 | 1.17e-16 | 1.17e-16 | 1.17e-16 |
| 0.005 | 0.0231 | 0.0238 | 1.17e-16 | 1.17e-16 | 1.17e-16 | 1.17e-16 | 1.17e-16 | 1.17e-16 | 1.17e-16 | 1.17e-16 |
| 0.1 | 0.0626 | 1.17e-16 | 1.17e-16 | 1.17e-16 | 1.17e-16 | 1.17e-16 | 1.17e-16 | 1.17e-16 | 1.17e-16 | 1.17e-16 |
| 0.5 | 1.17e-16 | 1.17e-16 | **1.17e-16** | 1.17e-16 | 1.17e-16 | 1.17e-16 | 1.17e-16 | 1.17e-16 | 1.17e-16 | 1.17e-16 |
| 0.7 | 1.17e-16 | 1.17e-16 | 1.17e-16 | 1.17e-16 | 1.17e-16 | 1.17e-16 | 1.17e-16 | 1.17e-16 | 1.17e-16 | 1.17e-16 |
| 1 | 1.17e-16 | 1.17e-16 | 1.17e-16 | 1.17e-16 | 1.17e-16 | 1.17e-16 | 1.17e-16 | 1.17e-16 | 1.17e-16 | 1.17e-16 |
| 1.5 | 1.17e-16 | 1.17e-16 | 1.17e-16 | 1.17e-16 | 1.17e-16 | 1.17e-16 | 1.17e-16 | 1.17e-16 | 1.17e-16 | 1.17e-16 |
| 10 | 1.17e-16 | 1.17e-16 | 1.17e-16 | 1.17e-16 | 1.17e-16 | 1.17e-16 | 1.17e-16 | 1.17e-16 | 1.17e-16 | 1.17e-16 |
| 50 | 1.17e-16 | 1.17e-16 | 1.17e-16 | 1.17e-16 | 1.17e-16 | 1.17e-16 | 1.17e-16 | 1.17e-16 | 1.17e-16 | 1.17e-16 |
| 100 | 1.17e-16 | 1.17e-16 | 1.17e-16 | 1.17e-16 | 1.17e-16 | 1.17e-16 | 1.17e-16 | 1.17e-16 | 1.17e-16 | 1.17e-16 |

TABLE III
MEANS AND STANDARD DEVIATIONS OF RI OBTAINED BY TFCM ON DATASET T1 IN TARGET DOMAIN BASED ON KNOWLEDGE OF DATASET S1-1 IN THE SOURCE DOMAIN UNDER DIFFERENT PARAMETER SETTINGS

| $\lambda_2$ \ $\lambda_1$ | 0 | 0.005 | 0.1 | 0.5 | 0.7 | 1 | 1.5 | 10 | 50 | 100 |
|---|---|---|---|---|---|---|---|---|---|---|
| Part A: Means | | | | | | | | | | |
| 0 | 0.8603 | 0.8623 | 0.8908 | 0.9081 | 0.8864 | 0.8706 | 0.8706 | 0.8706 | 0.8706 | 0.8706 |
| 0.005 | 0.8668 | 0.8659 | 0.8908 | 0.9103 | 0.8864 | 0.8706 | 0.8706 | 0.8706 | 0.8706 | 0.8706 |
| 0.1 | 0.8783 | 0.8908 | 0.9114 | 0.9109 | 0.8873 | 0.8706 | 0.8706 | 0.8706 | 0.8706 | 0.8706 |
| 0.5 | 0.9077 | 0.9133 | **0.9187** | 0.8873 | 0.8873 | 0.8706 | 0.8706 | 0.8706 | 0.8706 | 0.8706 |
| 0.7 | 0.8864 | 0.8864 | 0.8873 | 0.8873 | 0.8873 | 0.8706 | 0.8706 | 0.8706 | 0.8706 | 0.8706 |
| 1 | 0.8706 | 0.8706 | 0.8706 | 0.8706 | 0.8706 | 0.8706 | 0.8706 | 0.8706 | 0.8706 | 0.8706 |
| 1.5 | 0.8706 | 0.8706 | 0.8706 | 0.8706 | 0.8706 | 0.8706 | 0.8706 | 0.8706 | 0.8706 | 0.8706 |
| 10 | 0.8706 | 0.8706 | 0.8706 | 0.8706 | 0.8706 | 0.8706 | 0.8706 | 0.8706 | 0.8706 | 0.8706 |
| 50 | 0.8706 | 0.8706 | 0.8706 | 0.8706 | 0.8706 | 0.8706 | 0.8706 | 0.8706 | 0.8706 | 0.8706 |
| 100 | 0.8706 | 0.8706 | 0.8706 | 0.8706 | 0.8706 | 0.8706 | 0.8706 | 0.8706 | 0.8706 | 0.8706 |
| Part B: Standard deviations | | | | | | | | | | |
| 0 | 0.0335 | 0.0124 | 0 | 1.17e-16 | 1.17e-16 | 1.17e-16 | 2.34e-16 | 2.34e-16 | 0 | 0 |
| 0.005 | 0.0131 | 0.0130 | 1.17e-16 | 1.17e-16 | 1.17e-16 | 1.17e-16 | 2.34e-16 | 2.34e-16 | 0 | 0 |
| 0.1 | 0.0396 | 1.17e-16 | 0 | 1.17e-16 | 1.17e-16 | 1.17e-16 | 2.34e-16 | 2.34e-16 | 0 | 0 |
| 0.5 | 2.34e-16 | 2.34e-16 | **0** | 1.17e-16 | 2.34e-16 | 2.34e-16 | 2.34e-16 | 0 | 0 | 0 |
| 0.7 | 2.34e-16 | 1.17e-16 | 1.17e | 2.34e-16 | 2.34e-16 | 1.17e-16 | 0 | 0 | 0 | 0 |
| 1 | 2.34e-16 | 2.34e-16 | 1.17e | 2.34e-16 | 1.17e-16 | 1.17e-16 | 0 | 0 | 0 | 0 |
| 1.5 | 2.34e-16 | 2.34e-16 | 2.34e-16 | 1.17e-16 | 1.17e-16 | 0 | 0 | 0 | 0 | 0 |
| 10 | 0 | 0 | 0 | 0 | 0 | 0 | 0 | 0 | 0 | 0 |
| 50 | 0 | 0 | 0 | 0 | 0 | 0 | 0 | 0 | 0 | 0 |
| 100 | 0 | 0 | 0 | 0 | 0 | 0 | 0 | 0 | 0 | 0 |

TABLE IV
MEANS AND STANDARD DEVIATIONS OF THE XB INDEX OBTAINED BY TFCM ON DATASET T1 IN TARGET DOMAIN BASED ON KNOWLEDGE OF DATASET S1-1 IN THE SOURCE DOMAIN UNDER DIFFERENT PARAMETER SETTINGS

| $\lambda_2$ \ $\lambda_1$ | 0 | 0.005 | 0.1 | 0.5 | 0.7 | 1 | 1.5 | 10 | 50 | 100 |
|---|---|---|---|---|---|---|---|---|---|---|
| Part A: Means | | | | | | | | | | |
| 0 | 0.1911 | 0.1957 | 0.1770 | 0.1705 | 0.1748 | 0.1774 | 0.1774 | 0.1774 | 0.1774 | 0.1774 |
| 0.005 | 0.1803 | 0.1822 | 0.1770 | 0.1683 | 0.1748 | 0.1774 | 0.1774 | 0.1774 | 0.1774 | 0.1774 |
| 0.1 | 0.1799 | 0.1770 | 0.1677 | 0.1671 | 0.1721 | 0.1774 | 0.1774 | 0.1774 | 0.1774 | 0.1774 |
| 0.5 | 0.1701 | 0.1576 | **0.1555** | 0.1721 | 0.1721 | 0.1774 | 0.1774 | 0.1774 | 0.1774 | 0.1774 |
| 0.7 | 0.1748 | 0.1748 | 0.1721 | 0.1721 | 0.1721 | 0.1774 | 0.1774 | 0.1774 | 0.1774 | 0.1774 |
| 1 | 0.1774 | 0.1774 | 0.1774 | 0.1774 | 0.1774 | 0.1774 | 0.1774 | 0.1774 | 0.1774 | 0.1774 |
| 1.5 | 0.1774 | 0.1774 | 0.1774 | 0.1774 | 0.1774 | 0.1774 | 0.1774 | 0.1774 | 0.1774 | 0.1774 |
| 10 | 0.1774 | 0.1774 | 0.1774 | 0.1774 | 0.1774 | 0.1774 | 0.1774 | 0.1774 | 0.1774 | 0.1774 |
| 50 | 0.1774 | 0.1774 | 0.1774 | 0.1774 | 0.1774 | 0.1774 | 0.1774 | 0.1774 | 0.1774 | 0.1774 |
| 100 | 0.1774 | 0.1774 | 0.1774 | 0.1774 | 0.1774 | 0.1774 | 0.1774 | 0.1774 | 0.1774 | 0.1774 |
| Part B: Standard deviations | | | | | | | | | | |
| 0 | 0.0513 | 0.0029 | 3.50e-08 | 3.37e-14 | 1.24e-16 | 4.03e-17 | 4.71e-17 | 2.92e-17 | 2.92e-17 | 2.92e-17 |
| 0.005 | 0.0015 | 0.0034 | 1.07e-07 | 1.59e-14 | 1.04e-17 | 2.92e-17 | 8.62e-17 | 2.44e-17 | 2.92e-17 | 2.92e-17 |
| 0.1 | 0.0377 | 1.54e-07 | 8.36e-10 | 4.70e-16 | 5.31e-17 | 9.25e-18 | 4.23e-17 | 4.03e-17 | 3.81e-17 | 2.92e-17 |
| 0.5 | 1.41e-13 | 7.63e-14 | **1.61e-15** | 2.92e-17 | 2.92e-17 | 2.92e-17 | 2.61e-17 | 2.92e-17 | 0 | 5.39e-17 |
| 0.7 | 6.14e-16 | 6.68e-16 | 8.57e-17 | 4.33e-17 | 2.92e-17 | 2.92e-17 | 2.92e-17 | 2.92e-17 | 3.70e-17 | 2.92e-17 |
| 1 | 4.03e-17 | 3.06e-17 | 4.71e-17 | 4.33e-17 | 4.62e-17 | 2.92e-17 | 2.92e-17 | 0 | 3.58e-17 | 2.92e-17 |
| 1.5 | 2.92e-17 | 3.20e-17 | 2.26e-17 | 3.81e-17 | 8.06e-17 | 0 | 2.92e-17 | 2.92e-17 | 2.92e-17 | 0 |
| 10 | 0 | 4.33e-17 | 0 | 1.85e-17 | 2.92e-17 | 0 | 2.92e-17 | 0 | 1.60e-17 | 2.92e-17 |
| 50 | 0 | 2.92e-17 | 0 | 0 | 2.92e-17 | 2.92e-17 | 2.92e-17 | 2.92e-17 | 0 | 2.92e-17 |
| 100 | 0 | 2.92e-17 | 2.92e-17 | 2.92e-17 | 2.92e-17 | 2.92e-17 | 2.92e-17 | 2.92e-17 | 2.92e-17 | 2.92e-17 |

TABLE V
PERFORMANCE (OPTIMIZED PARAMETERS WITH THE XB INDEX) COMPARISON OF FCM AND TFCM ON SYNTHETIC DATASET T1 WITH THE DATASET S1-1 AS THE SOURCE DATASET.

| Indices | Source domain dataset S1-1 | | | | Target domain dataset T1 | | | | | | | |
|---|---|---|---|---|---|---|---|---|---|---|---|---|
| | FCM | | | | FCM | | | | TFCM (S1-1=>T1)* | | | |
| | min | max | mean | std | min | max | mean | std | min | max | mean | std |
| *NMI* | 0.8743 | 0.8743 | 0.8743 | 9.06e-017 | 0.5032 | 0.6730 | 0.6561 | 0.0536 | 0.7850 | 0.7850 | **0.7850** | **1.17e-16** |
| *RI* | 0.9651 | 0.9651 | 0.9651 | 0 | 0.7648 | 0.8708 | 0.8603 | 0.0335 | 0.9187 | 0.9187 | **0.9187** | **0** |
| *XB* | 0.0742 | 0.0742 | 0.0742 | 1.01e-007 | 0.1748 | 0.3371 | 0.1911 | 0.0513 | 0.1555 | 0.1555 | **0.1555** | **1.61e-15** |

* For a $\Longrightarrow$ b, b is the dataset in the target domain and a is the dataset in the corresponding source domain.

TABLE VI
MEANS AND STANDARD DEVIATIONS OF NMI OBTAINED BY TFCM ON DATASET T1 IN TARGET DOMAIN BASED ON KNOWLEDGE OF DATASET S1-2 IN THE SOURCE DOMAIN UNDER DIFFERENT PARAMETER SETTINGS

| $\lambda_1$ / $\lambda_2$ | 0 | 0.005 | 0.1 | 0.5 | 0.7 | 1 | 1.5 | 10 | 50 | 100 |
|---|---|---|---|---|---|---|---|---|---|---|
| Part A: Means | | | | | | | | | | |
| 0 | **0.6603** | 0.6348 | 0.6354 | 0.6251 | 0.6251 | 0.5301 | 0.5301 | 0.5301 | 0.5301 | 0.5301 |
| 0.005 | 0.6358 | 0.6321 | 0.6354 | 0.6251 | 0.6251 | 0.5301 | 0.5301 | 0.5301 | 0.5301 | 0.5301 |
| 0.1 | 0.6354 | 0.6354 | 0.6354 | 0.6251 | 0.5598 | 0.5301 | 0.5301 | 0.5301 | 0.5301 | 0.5301 |
| 0.5 | 0.6251 | 0.6251 | 0.6251 | 0.5301 | 0.5301 | 0.5301 | 0.5301 | 0.5301 | 0.5301 | 0.5301 |
| 0.7 | 0.6251 | 0.6251 | 0.5598 | 0.5301 | 0.5301 | 0.5301 | 0.5301 | 0.5301 | 0.5301 | 0.5301 |
| 1 | 0.5301 | 0.5301 | 0.5301 | 0.5301 | 0.5301 | 0.5301 | 0.5301 | 0.5301 | 0.5301 | 0.5301 |
| 1.5 | 0.5301 | 0.5301 | 0.5301 | 0.5301 | 0.5301 | 0.5301 | 0.5301 | 0.5301 | 0.5301 | 0.5301 |
| 10 | 0.5301 | 0.5301 | 0.5301 | 0.5301 | 0.5301 | 0.5301 | 0.5301 | 0.5301 | 0.5301 | 0.5301 |
| 50 | 0.5301 | 0.5301 | 0.5301 | 0.5301 | 0.5301 | 0.5301 | 0.5301 | 0.5301 | 0.5301 | 0.5301 |
| 100 | 0.5301 | 0.5301 | 0.5301 | 0.5301 | 0.5301 | 0.5301 | 0.5301 | 0.5301 | 0.5301 | 0.5301 |
| Part B: Standard deviations | | | | | | | | | | |
| 0 | 0.0447 | 0.0254 | 0 | 1.17e-16 | 1.17e-16 | 1.17e-16 | 1.17e-16 | 1.17e-16 | 1.17e-16 | 1.17e-16 |
| 0.005 | 0.0222 | 0.0457 | 0 | 1.17e-16 | 1.17e-16 | 1.17e-16 | 1.17e-16 | 1.17e-16 | 1.17e-16 | 1.17e-16 |
| 0.1 | 0 | 0 | 0 | 1.17e-16 | 0 | 1.17e-16 | 1.17e-16 | 1.17e-16 | 1.17e-16 | 1.17e-16 |
| 0.5 | 1.17e-16 | 1.17e-16 | 1.17e-16 | 1.17e-16 | 1.17e-16 | 1.17e-16 | 1.17e-16 | 1.17e-16 | 1.17e-16 | 1.17e-16 |
| 0.7 | 1.17e-16 | 1.17e-16 | 0 | 1.17e-16 | 1.17e-16 | 1.17e-16 | 1.17e-16 | 1.17e-16 | 1.17e-16 | 1.17e-16 |
| 1 | 1.17e-16 | 1.17e-16 | 1.17e-16 | 1.17e-16 | 1.17e-16 | 1.17e-16 | 1.17e-16 | 1.17e-16 | 1.17e-16 | 1.17e-16 |
| 1.5 | 1.17e-16 | 1.17e-16 | 1.17e-16 | 1.17e-16 | 1.17e-16 | 1.17e-16 | 1.17e-16 | 1.17e-16 | 1.17e-16 | 1.17e-16 |
| 10 | 1.17e-16 | 1.17e-16 | 1.17e-16 | 1.17e-16 | 1.17e-16 | 1.17e-16 | 1.17e-16 | 1.17e-16 | 1.17e-16 | 1.17e-16 |
| 50 | 1.17e-16 | 1.17e-16 | 1.17e-16 | 1.17e-16 | 1.17e-16 | 1.17e-16 | 1.17e-16 | 1.17e-16 | 1.17e-16 | 1.17e-16 |
| 100 | 1.17e-16 | 1.17e-16 | 1.17e-16 | 1.17e-16 | 1.17e-16 | 1.17e-16 | 1.17e-16 | 1.17e-16 | 1.17e-16 | 1.17e-16 |

TABLE VII
MEANS AND STANDARD DEVIATIONS OF RI OBTAINED BY TFCM ON DATASET T1 IN TARGET DOMAIN BASED ON KNOWLEDGE OF DATASET S1-2 IN THE SOURCE DOMAIN UNDER DIFFERENT PARAMETER SETTINGS

| $\lambda_1$ / $\lambda_2$ | 0 | 0.005 | 0.1 | 0.5 | 0.7 | 1 | 1.5 | 10 | 50 | 100 |
|---|---|---|---|---|---|---|---|---|---|---|
| Part A: Means | | | | | | | | | | |
| 0 | **0.8605** | 0.8502 | 0.8491 | 0.8427 | 0.8427 | 0.7858 | 0.7858 | 0.7858 | 0.7858 | 0.7858 |
| 0.005 | 0.8503 | 0.8504 | 0.8491 | 0.8427 | 0.8427 | 0.7858 | 0.7858 | 0.7858 | 0.7858 | 0.7858 |
| 0.1 | 0.8491 | 0.8491 | 0.8491 | 0.8427 | 0.8028 | 0.7858 | 0.7858 | 0.7858 | 0.7858 | 0.7858 |
| 0.5 | 0.8427 | 0.8427 | 0.8427 | 0.7858 | 0.7858 | 0.7858 | 0.7858 | 0.7858 | 0.7858 | 0.7858 |
| 0.7 | 0.8427 | 0.8427 | 0.8028 | 0.7858 | 0.7858 | 0.7858 | 0.7858 | 0.7858 | 0.7858 | 0.7858 |
| 1 | 0.7858 | 0.7858 | 0.7858 | 0.7858 | 0.7858 | 0.7858 | 0.7858 | 0.7858 | 0.7858 | 0.7858 |
| 1.5 | 0.7858 | 0.7858 | 0.7858 | 0.7858 | 0.7858 | 0.7858 | 0.7858 | 0.7858 | 0.7858 | 0.7858 |
| 10 | 0.7858 | 0.7858 | 0.7858 | 0.7858 | 0.7858 | 0.7858 | 0.7858 | 0.7858 | 0.7858 | 0.7858 |
| 50 | 0.7858 | 0.7858 | 0.7858 | 0.7858 | 0.7858 | 0.7858 | 0.7858 | 0.7858 | 0.7858 | 0.7858 |
| 100 | 0.7858 | 0.7858 | 0.7858 | 0.7858 | 0.7858 | 0.7858 | 0.7858 | 0.7858 | 0.7858 | 0.7858 |
| Part B: Standard deviations | | | | | | | | | | |
| 0 | 0.0327 | 0.0145 | 1.17e-16 | 0 | 0 | 0 | 0 | 0 | 0 | 0 |
| 0.005 | 0.0127 | 0.0269 | 1.17e-16 | 0 | 0 | 0 | 0 | 0 | 0 | 0 |
| 0.1 | 1.17e-16 | 1.17e-16 | 1.17e-16 | 0 | 1.17e-16 | 0 | 0 | 0 | 0 | 0 |
| 0.5 | 0 | 0 | 0 | 0 | 0 | 0 | 0 | 0 | 0 | 0 |
| 0.7 | 0 | 0 | 1.17e-16 | 0 | 0 | 0 | 0 | 0 | 0 | 0 |
| 1 | 0 | 0 | 0 | 0 | 0 | 0 | 0 | 0 | 0 | 0 |
| 1.5 | 0 | 0 | 0 | 0 | 0 | 0 | 0 | 0 | 0 | 0 |
| 10 | 0 | 0 | 0 | 0 | 0 | 0 | 0 | 0 | 0 | 0 |
| 50 | 0 | 0 | 0 | 0 | 0 | 0 | 0 | 0 | 0 | 0 |
| 100 | 0 | 0 | 0 | 0 | 0 | 0 | 0 | 0 | 0 | 0 |

TABLE VIII
MEANS AND STANDARD DEVIATIONS OF THE XB INDEX OBTAINED BY TFCM ON DATASET T1 IN TARGET DOMAIN BASED ON KNOWLEDGE OF DATASET S1-2 IN THE SOURCE DOMAIN UNDER DIFFERENT PARAMETER SETTINGS

| $\lambda_2$ \ $\lambda_1$ | 0 | 0.005 | 0.1 | 0.5 | 0.7 | 1 | 1.5 | 10 | 50 | 100 |
|---|---|---|---|---|---|---|---|---|---|---|
| Part A: Means | | | | | | | | | | |
| 0 | **0.1928** | 0.2095 | 0.1979 | 0.2000 | 0.2000 | 0.2455 | 0.2455 | 0.2455 | 0.2455 | 0.2455 |
| 0.005 | 0.2062 | 0.2061 | 0.1979 | 0.2000 | 0.2000 | 0.2455 | 0.2455 | 0.2455 | 0.2455 | 0.2455 |
| 0.1 | 0.1979 | 0.1979 | 0.1979 | 0.2000 | 0.2219 | 0.2455 | 0.2455 | 0.2455 | 0.2455 | 0.2455 |
| 0.5 | 0.2000 | 0.2000 | 0.2000 | 0.2455 | 0.2455 | 0.2455 | 0.2455 | 0.2455 | 0.2455 | 0.2455 |
| 0.7 | 0.2000 | 0.2000 | 0.2219 | 0.2455 | 0.2455 | 0.2455 | 0.2455 | 0.2455 | 0.2455 | 0.2455 |
| 1 | 0.2455 | 0.2455 | 0.2455 | 0.2455 | 0.2455 | 0.2455 | 0.2455 | 0.2455 | 0.2455 | 0.2455 |
| 1.5 | 0.2455 | 0.2455 | 0.2455 | 0.2455 | 0.2455 | 0.2455 | 0.2455 | 0.2455 | 0.2455 | 0.2455 |
| 10 | 0.2455 | 0.2455 | 0.2455 | 0.2455 | 0.2455 | 0.2455 | 0.2455 | 0.2455 | 0.2455 | 0.2455 |
| 50 | 0.2455 | 0.2455 | 0.2455 | 0.2455 | 0.2455 | 0.2455 | 0.2455 | 0.2455 | 0.2455 | 0.2455 |
| 100 | 0.2455 | 0.2455 | 0.2455 | 0.2455 | 0.2455 | 0.2455 | 0.2455 | 0.2455 | 0.2455 | 0.2455 |
| Part B: Standard deviations | | | | | | | | | | |
| 0 | 0.0563 | 0.0503 | 6.08e-07 | 1.34e-16 | 7.28e-17 | 3.92e-17 | 4.62e-17 | 7.16e-17 | 1.05e-16 | 1.43e-16 |
| 0.005 | 0.0452 | 0.0454 | 1.06e-09 | 1.96e-16 | 6.60e-17 | 2.92e-17 | 2.92e-17 | 7.16e-17 | 1.60e-17 | 8.06e-17 |
| 0.1 | 1.47e-09 | 6.16e-07 | 5.66e-12 | 5.31e-17 | 3.20e-17 | 4.62e-17 | 2.92e-17 | 5.85e-17 | 2.92e-17 | 0 |
| 0.5 | 6.46e-16 | 9.79e-16 | 5.47e-17 | 4.13e-17 | 4.33e-17 | 0 | 0 | 5.85e-17 | 0 | 0 |
| 0.7 | 4.80e-17 | 6.67e-17 | 3.92e-17 | 3.20e-17 | 4.33e-17 | 2.92e-17 | 2.61e-17 | 2.92e-17 | 4.62e-17 | 5.85e-17 |
| 1 | 2.06e-17 | 5.39e-17 | 2.61e-17 | 0 | 1.85e-17 | 2.92e-17 | 2.92e-17 | 5.85e-17 | 5.85e-17 | 0 |
| 1.5 | 2.92e-17 | 2.92e-17 | 0 | 2.61e-17 | 2.92e-17 | 2.92e-17 | 5.39e-17 | 6.54e-17 | 0 | 0 |
| 10 | 0 | 2.92e-17 | 5.99e-17 | 2.92e-17 | 0 | 2.92e-17 | 4.03e-17 | 0 | 0 | 5.85e-17 |
| 50 | 2.92e-17 | 0 | 0 | 2.92e-17 | 2.92e-17 | 2.92e-17 | 2.92e-17 | 2.92e-17 | 2.92e-17 | 2.92e-17 |
| 100 | 0 | 2.92e-17 | 2.92e-17 | 0 | 0 | 0 | 2.77e-17 | 5.85e-17 | 2.92e-17 | 0 |

TABLE IX
PERFORMANCE (OPTIMIZED PARAMETERS WITH XB INDEX) COMPARISON OF FCM AND TFCM ON SYNTHETIC DATASET T1 WITH THE DATASET S1-2 AS THE SOURCE DATASET.

| Indices | Source domain dataset S1-2 | | | | Target domain dataset T1 | | | | | | | |
|---|---|---|---|---|---|---|---|---|---|---|---|---|
| | FCM | | | | FCM | | | | TFCM (S1-2=>T1)* | | | |
| | min | max | mean | std | min | max | mean | std | min | max | mean | std |
| *NMI* | 0.8620 | 0.8620 | 0.8620 | 1.17e-16 | 0.5032 | 0.6730 | 0.6561 | 0.0536 | 0.5314 | 0.6730 | **0.6603** | 0.0447 |
| *RI* | 0.9609 | 0.9609 | 0.9609 | 0 | 0.7648 | 0.8708 | 0.8603 | 0.0335 | 0.7674 | 0.8708 | **0.8605** | 0.0327 |
| *XB* | 0.0856 | 0.0856 | 0.0856 | 5.99e-08 | 0.1748 | 0.3371 | 0.1911 | 0.0513 | 0.1749 | 0.3530 | **0.1928** | 0.0563 |

* For a $\Rightarrow$ b, b is the dataset in the target domain and a is the dataset in the corresponding source domain.

*2) How to Control the Negative Transfer from Bad Source Domain*

Another important concern of the proposed algorithms is that if the source domain contains bad information, how to effectively control the negative influence of the bad information. For this issue, we still take TFCM as the example to give an analysis based on the dataset T1 in the target domain and the dataset S1-2 in a bad source domain, as shown in Table 1 and Fig. 4.

Tables VI-VIII record the results of TFCM on the target dataset T1 with the dataset S1-2 as the source domain dataset. Based on these results, the following observations can be obtained.

(1) When the two parameters in Eq.(5) are set as $\lambda_1 = 0$ and $\lambda_2 = 0$, TFCM obtained the best clustering effect as indicated by the mean *NMI*, *RI* and *XB* values in Part A of Tables VI-VIII, while the clustering performance has been further degenerated when $\lambda_1$ and $\lambda_2$ are set to different positive values. Here, when $\lambda_1 = 0$ and $\lambda_2 = 0$, TFCM is the same as the classical FCM. Thus the results in Part A of Tables III and IV show that the bad knowledge about the source domain indeed has a negative influence on the clustering performance in the target domain.

(2) From the above analysis, we know that it is very necessary to control the negative influence of the bad source domain on the target domain, which actually means how to weaken the influence of the bad source domain by adopting the appropriate parameter values. Thus, the *XB* and $XB_{FSC}$ indices can be used again as the criteria to determine the appropriate parameter values for TFCM/E-TFCM and TFSC/E-TFSC, respectively, to control the negative influence of the bad source domain.

In Table IX, the optimized results of TFCM on the dataset T1 with the dataset S1-2 as the source domain dataset are reported, where the parameters ( $\lambda_1 = 0$ , $\lambda_2 = 0$ ,) were determined with the *XB* index as the evaluation criterion. It shows that the optimized clustering results of TFCM are almost the same to that of the classical FCM algorithm on dataset T1. The slight difference of clustering results between FCM and TFCM ($\lambda_1 = 0, \lambda_2 = 0$,) on T1 is caused by the different initializations of 20 runs.

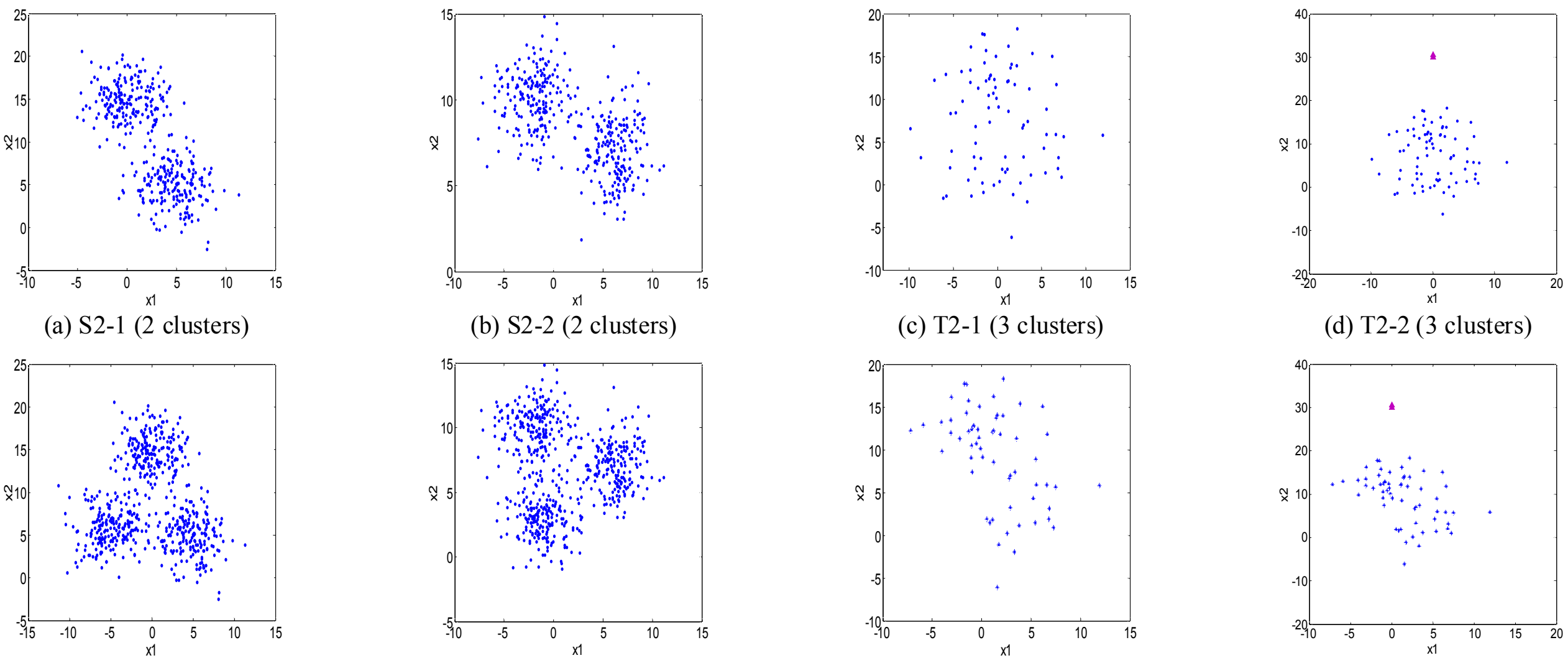

(a) S2-1 (2 clusters) (b) S2-2 (2 clusters) (c) T2-1 (3 clusters) (d) T2-2 (3 clusters)

(e) S3-1 (3 clusters) (f) S3-2 (3 clusters) (g) T3-1 (2 clusters) (h) T3-2 (2 clusters)

Fig. 5 Synthetic datasets used for the knowledge transfer where the numbers of clusters are different between the source domain and the target domain. S2-1 and S2-2 in (a) and (b) are used as the good and bad source domain datasets for the target domain datasetsT2-1 and T2-2 in (c) and (d); S3-1 and S3-2 in (e) and (f) are used as the good and bad source domain datasets for the target domain datasets T3-1 and T3-2 in (g) and (h).

TABLE X

PERFORMANCE EVALUATION ON THE DATASETS WITH DIFFERENT NUMBERS OF CLUSTERS BETWEEN THE SOURCE AND TARGET DOMAINS

| Dataset T2-1 | | | | | | |
|---|---|---|---|---|---|---|
| Indices | FCM | | E-TFCM | | | |
| | | | S2-1 (2 clusters) ⟹ T2-1 (3 clusters) (Good source) | | S2-2 (2 clusters) ⟹ T2-1 (3 clusters) * (Bad source) | |
| | mean | std | mean | std | mean | std |
| *NMI* | 0.6561 | 0.0536 | **0.7627** | **0** | 0.6572 | 0.0386 |
| *RI* | 0.8603 | 0.0335 | **0.9158** | **0** | 0.8605 | 0.0258 |
| *XB* | 0.1911 | 0.0513 | **0.1687** | **0** | 0.1873 | 0.0392 |
| Optimized parameters | | | $\lambda_1 = 0.1$ | | $\lambda_1 = 0$ | |
| Dataset T2-2 | | | | | | |
| Indices | FCM | | E-TFCM | | | |
| | | | S2-1 (2 clusters) ⟹ T2-2 (3 clusters) (Good source) | | S2-2 (2 clusters) ⟹ T2-2 (3 clusters) (Bad source) | |
| | mean | std | mean | std | mean | std |
| *NMI* | 0.4955 | 0.0530 | **0.6612** | **0** | 0.4870 | 0.0617 |
| *RI* | 0.7746 | 0.0402 | **0.8750** | **1.17e-16** | 0.7641 | 0.0444 |
| *XB* | 0.2869 | 0.0954 | **0.1906** | **2.95e-08** | 0.3089 | 0.1068 |
| Optimized parameters | | | $\lambda_1 = 0.1$ | | $\lambda_1 = 0$ | |
| Dataset T3-1 | | | | | | |
| Indices | FCM | | E-TFCM | | | |
| | | | S3-1 (3 clusters) ⟹ T3-1 (2 clusters) (Good source) | | S3-2 (3 clusters) ⟹ T3-1 (2 clusters) (Bad source) | |
| | mean | std | mean | std | mean | std |
| *NMI* | 0.7882 | 1.17e-16 | **0.8114** | **0** | 0.7882 | 1.17e-16 |
| *RI* | 0.9145 | 1.17e-16 | **0.9344** | **0** | 0.9145 | 1.17e-16 |
| *XB* | 0.2024 | 2.27e-07 | **0.1361** | **0** | 0.2024 | 2.27e-07 |
| Optimized parameters | | | $\lambda_1 = 0.5$ | | $\lambda_1 = 0$ | |
| Dataset T3-2 | | | | | | |
| Indices | FCM | | E-TFCM | | | |
| | | | S3-1 (3 clusters) ⟹ T3-2 (2 clusters) (Good source) | | S3-2 (3 clusters) ⟹ T3-2 (2 clusters) (Bad source) | |
| | mean | std | mean | std | mean | std |
| *NMI* | 0.7014 | 0 | **0.8142** | **0** | 0.7014 | 0 |
| *RI* | 0.8773 | 0 | **0.9365** | **1.24e-16** | 0.8773 | 0 |
| *XB* | 0.3614 | 3.92e-17 | **0.1331** | **0** | 0.3614 | 3.92e-17 |
| Optimized parameters | | | $\lambda_1 = 0.7$ | | $\lambda_1 = 0$ | |

* For a ⟹ b, b is the dataset in the target domain and a is the dataset in the corresponding source domain.

*3) Different Numbers of Clusters between the Source and Target Domains*

In this subsection, the performance is evaluated in the case where the numbers of clusters of the source domain and target domain are different. Here, we take the E-TFCM algorithm as an example. The results of the proposed E-TFSC algorithm will reported in subsection IV-D for the comprehensive comparison. Synthetic datasets used are given in Fig. 5, where eight datasets are denoted as S2-1, S2-2, T2-1, T2-2, S3-1, S3-2, T3-1, T3-2, respectively. S2-1 and S2-2 in Fig. 5(a) and 5(b) are used as the good and bad source domain datasets for the target domain datasets T2-1 and T2-2 in Fig. 5(c) and 5(d); S3-1 and S3-2 in Fig. 5(e) and (f) are used as the good and bad source domain datasets for the target domain datasets T3-1 and T3-2 in Fig. 5(g) and (h). Here, T2-2 and T3-2 are generated based on T2-1 and T3-1 with a few outlier added. Finally, eight datasets pairs are used to evaluate the performance of the E-TFCM algorithm.

The clustering results are reported in Table X. From these results we can get the following observations.

(1) Although the numbers of clusters between the source domain and the target domain are different, the transfer learning still can be implemented by the E-TFCM algorithm.

(2) When the source domain consists of the useful information for the target domain, the clustering performance can be effectively enhanced with these information by knowledge transfer.

(3) When the source domain consisting of the bad information for the target domain, the negative influence of clustering performance can also be controlled effectively by the optimized parameters determined with the XB index.

*4) High Dimensional Datasets*

In this study, TFSC and E-TFSC are presented due to the fact that TFCM and E-TFCM are not very effective for high dimensional datasets. In this subsection, two synthetic datasets, denoted as S4 and T4, with the predetermined cluster structures are used to evaluate the performance of TFSC on the high dimensional data. The results of E-TFSC will be reported in section IV-D for the comprehensive performance comparison. Datasets are generated for the source domain and the target domain, respectively. Each domain contains three clusters located at different subspace as shown in Fig 6. The corresponding parameters used to generate the data are listed in Table XI. Fig. 6(a) and Fig. 6(b) contain three sub-figures, each of which corresponds to a cluster where the high dimensional data in this cluster are taken as the sequences and plotted in the corresponding sub-figure. In the sub-figures, the sequence number of features and the value of features are taken as the x-coordinate and y-coordinate, respectively. From these sub-figures, we can see the corresponding subspace of each cluster, in which the features are more important than the other features to the associated cluster. For example, from three sub-figures in Fig. 6(a), we can see that most features of three clusters have the feature values uniformly distributed in the interval [0, 100], but for each cluster there is a subset of features whose values are distinctive from other clusters. As shown in Fig. 6(a), three clusters in the source domain have the distinctive feature subsets with the sequence number of features in the intervals of [1, 31], [10, 40] and [20, 55], respectively.

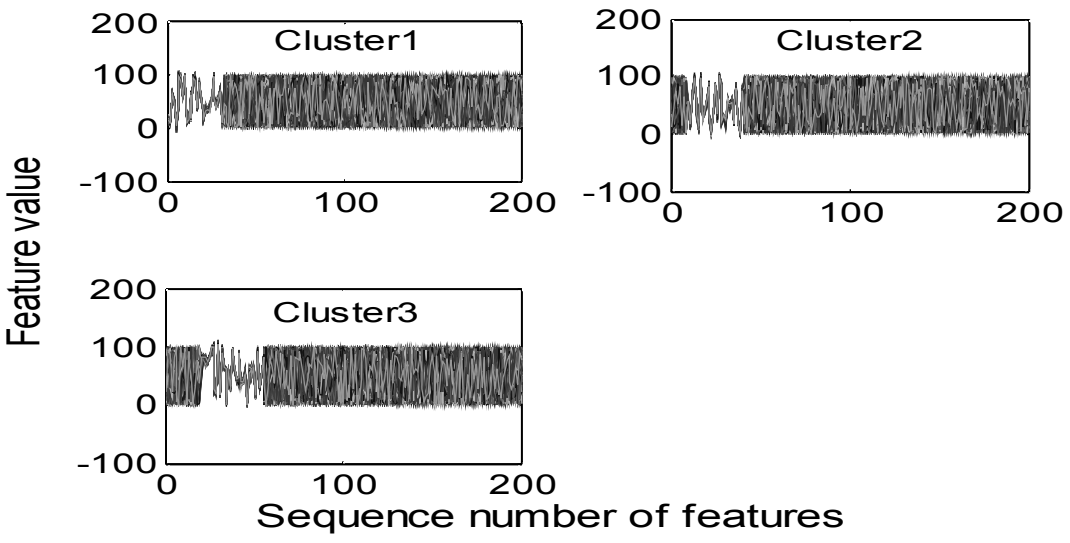

(a) Source domain S4

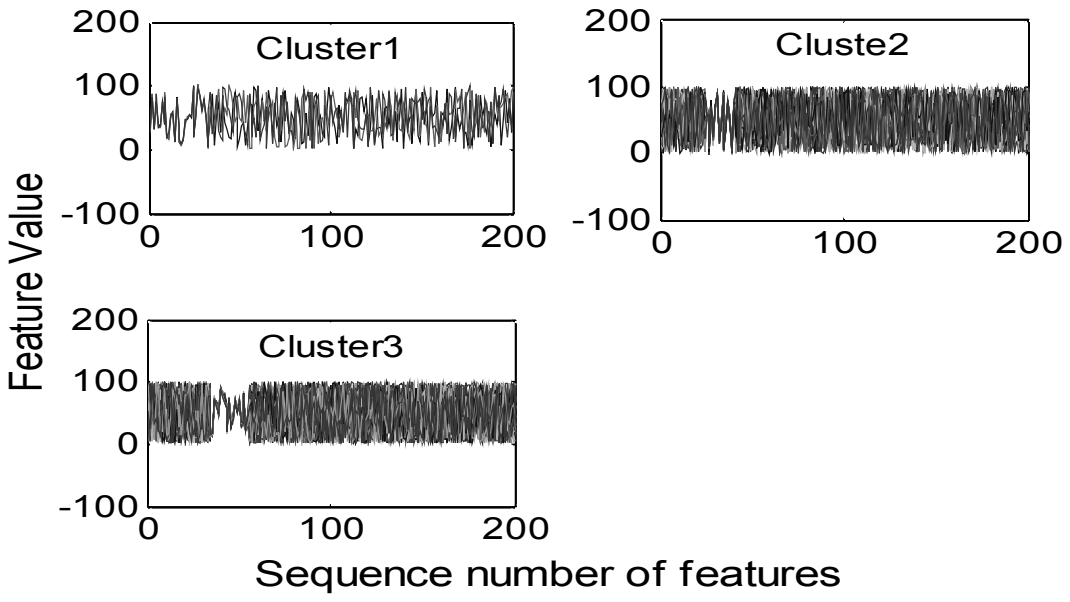

(b) Target domain T4

Fig.6 Synthetic dataset generated for evaluating the proposed TFSC algorithm: (a) data generated in the source domain (S4) ; (b) data generated in the target domain (T4)

TABLE XI
PARAMETERS USED TO GENERATE SYNTHETIC DATASETS4 AND T4 FOR EVALUATING THE TFSC ALGORITHM

| Domain | | Source domain S4 | Target domain T4 |
|---|---|---|---|
| Important attribute set | Cluster-1 | [1:31] | [1:32] |
| | Cluster-2 | [10:40] | [25:40] |
| | Cluster-3 | [20:55] | [35:55] |
| size | | 600 | 60 |

The performance of TFSC is compared with that of the classical FSC algorithm with the results under the optimized parameters reported in Table XII. Here the optimized parameters are determined by the improve *XB* index, i.e., $XB_{FSC}$. Basically, the observations for TFSC are very similar to those for TFCM. Let's summarize them as follows. Table XII shows that the clustering result of TFSC with optimized parameter are obviously superior to that of the classical FSC algorithm. In particular, TFSC has obtained the mean values of both *NMI* and *RI* as 1 for the data in the target domain, which implies that all the data have been correctly clustered, while the clustering effect of FSC is inferior to that of TFSC obviously.

In addition to the aforementioned experimental results, the subspace identification ability of TFSC is also compared with that of FSC. The results are reported in Table XIII with the identification rates computed. Obviously, the identification rate of FSC (35/69) has been improved to that of TFSC (52/69) by leveraging the knowledge of the source domain. Specifically, Table XIII shows that TFSC has identified more truly important features than FSC in the corresponding embedded subspace for each cluster. For example, TFSC has detected 24 important features while FSC only identifies 9 important features for the 1st cluster.

TABLE XII

PERFORMANCE (OPTIMIZED PARAMETERS) COMPARISON OF FSC AND TFSC ON SYNTHETIC DATASET T4 WITH S4 AS THE DATASET IN THE SOURCE DOMAIN.

| Indices | FSC | | | | | | | | TFSC | | | |
|---|---|---|---|---|---|---|---|---|---|---|---|---|
| | Source domain dataset S4 | | | | Target domain dataset T4 | | | | Target domain dataset T4 (S4=>T4)* | | | |
| | min | max | mean | std | min | max | mean | std | min | max | mean | std |
| NMI | 0.9943 | 0.9943 | 0.9943 | 0 | 0.5434 | 0.8163 | 0.7084 | 0.0990 | 1 | 1 | **1** | 1.17e-016 |
| RI | 0.9999 | 0.9999 | 0.9999 | 0 | 0.7386 | 0.8961 | 0.8220 | 0.0553 | 1 | 1 | **1** | 0 |
| $XB_{FSC}$ | 0.1298 | 0.1298 | 0.1298 | 0 | 2.3113 | 2.5413 | 2.3418 | 0.0702 | 0.4221 | 0.4221 | **0.4221** | 2.89e-05 |

* For a $\Rightarrow$ b, b is the dataset in the target domain and a is the dataset in the corresponding source domain.

TABLE XIII

COMPARISON OF IDENTIFICATION ABILITIES OF FSC AND TFSC ON DATASET T4.

| No. of cluster | Sequence no. of important features | | |
|---|---|---|---|
| | True | Identified by FSC | Identified by TFSC |
| 1 | 32 features: {1:32} | 32 features with top weights: {37,51,36,**26***,**20**,38,179,43,25,191,42,141,104,183,**12**,41,**23**,**11**, 39,144,**16**,130,146,28,**33**,**31**,101,143,46,**17**,100,126} | 32 features with top weights: {**12**,**16**,**27**,**13**,98,63,156,**8**,**5**,144,162,**6**,183,**3**,106,160,**1**,**23**,**30**,**24**, **7**,**22**,**10**,**4**,**11**,**21**,**15**,**31**,**20**,**17**,**14**,**2**} |
| 2 | 16 features: {25:40} | 16 features with top weights: {**37**,51,**36**,**26**,20,**38**,179,43,**25**,191,42,141,104,183,12,41} | 16 features with top weights: {**37**,**33**,**40**,**39**,**29**,**36**,**35**,**28**,148,155,132,175,63,115,142,74} |
| 3 | 21 features: {35:55} | 21 features with top weights: {**35**,**37**,**38**,**40**,**36**,**52**,**49**,**43**,**51**,**50**,**46**,**48**,**42**,**41**,**45**,**47**,**53**,**54**,**39**,**55**,**44**} | 21 features with top weights: {**52**,**49**,**38**,**43**,**39**,**51**,**47**,**55**,**48**,**37**,**45**,**50**,**46**,**42**,**44**,**54**,**53**,**41**,**36**,**35**,65} |
| Identification Rate | | (9+5+21)/69=35/69 | (24+8+20)/69=52/69 |

* The bold number denotes the detected truly important features associated with the embedded subspaces.

TABLE XV

FOUR PAIRS OF 20 NEWSGROUPS TEXT DATASETS CONSTRUCTED FOR PERFORMANCE EVALUATION

| Dataset Pair | Source domain | | | Target domain | | |
|---|---|---|---|---|---|---|
| | Dataset | Number of clusters | Clusters and subcluster | Dataset | Number of clusters | Clusters and subcluster |
| 1 | NG20-S1 | 3 | *comp* (comp.graphics<br>comp.os.ms-windows.misc<br>comp.sys.ibm.pc.hardware<br>comp.sys.mac.hardware) | NG20-T1 | 3 | *comp.windows.x* |
| | | | *rec* (rec.autos<br>rec.motorcycles<br>rec.sport.baseball) | | | *rec.sport.hockey* |
| | | | *sci* (sci.crypt<br>sci.electronics<br>sci.med) | | | *sci.space* |
| | Size | | *3982*+2993+2995 | Size | | 998+998+999 |
| 2 | NG20-S2 | 3 | *comp* (comp.graphics<br>comp.os.ms-windows.misc<br>comp.sys.ibm.pc.hardware<br>comp.sys.mac.hardware) | NG20-T1 | 3 | *comp.windows.x* |
| | | | *rec* (rec.autos<br>rec.motorcycles<br>rec.sport.baseball) | | | *rec.sport.hockey* |
| | | | *talk* (talk.politics.uns<br>talk.politics.mideast<br>talk.politics.misc) | | | *sci.space* |
| | Size | | 3982+2993+2998 | Size | | 998+998+999 |
| 3 | NG20-S2 | 3 | *comp* (comp.graphics<br>comp.os.ms-windows.misc<br>comp.sys.ibm.pc.hardware<br>comp.sys.mac.hardware) | NG20-T2 | 2 | *comp.windows.x* |
| | | | *rec* (rec.autos<br>rec.motorcycles<br>rec.sport.baseball) | | | *rec.sport.hockey* |
| | | | *talk* (talk.politics.uns<br>talk.politics.mideast<br>talk.politics.misc) | | | |
| | Size | | 3982+2993+2998 | Size | | 998+998 |
| 4 | NG20-S3 | 2 | *comp* (comp.graphics<br>comp.os.ms-windows.misc<br>comp.sys.ibm.pc.hardware<br>comp.sys.mac.hardware) | NG20-T1 | 3 | *comp.windows.x* |
| | | | *talk* (talk.politics.uns<br>talk.politics.mideast<br>talk.politics.misc) | | | *rec.sport.hockey* |
| | | | | | | *sci.space* |
| | Size | | 3982+2998 | Size | | 998+998+999 |

TABLE XVI
PERFORMANCE COMPARISON BETWEEN THE CLASSICAL CLUSTERING ALGORITHMS AND THE TRANSFER PROTOTYPE-BASED CLUSTERING ALGORITHMS ON TWO 20NG TEXT DATASETS WITH THE SAME NUMBER OF CLUSTERS BETWEEN SOURCE DOMAIN AND TARGET DOMAIN

| Datasets | Indices | FCM | TFCM | E-TFCM | FSC | TFSC | E-TFSC |
|---|---|---|---|---|---|---|---|
| NG20-T1 (NG20-S1 ⇒ NG20-T1) | NMI-mean | 0.0184 | 0.0314 | 0.0247 | 0.5561 | **0.6711** | 0.6394 |
| | NMI-std | 0 | 0 | 0 | 0.0135 | **0.0056** | 0.0157 |
| | RI-mean | 0.2271 | 0.3369 | 0.3178 | 0.7573 | **0.8462** | 0.7972 |
| | RI-std | 0 | 0 | 0 | 0.0215 | **0.0088** | 0.0464 |
| | $XB\backslash XB_{FSC}$ -mean | 0.0384 | 0.0341 | 0.0362 | 0.0002 | **1.53e-05** | 0.0001 |
| | $XB\backslash XB_{FSC}$ -std | 9.4e-006 | 8.0e-17 | 1.19e-08 | 9.97e-05 | **1.70e-06** | 2.97e-05 |
| NG20-T1 (NG20-S2 ⇒ NG20-T1) | NMI-mean | 0.0184 | 0.0271 | 0.0196 | 0.5561 | **0.6383** | 0.6270 |
| | NMI-std | 0 | 0 | 0 | 0.0135 | **0.0139** | 0.0266 |
| | RI-mean | 0.2271 | 0.3273 | 0.2886 | 0.7573 | **0.8065** | 0.8067 |
| | RI-std | 0 | 0 | 0 | 0.0215 | **0.0069** | 0.0172 |
| | $XB\backslash XB_{FSC}$ -mean | 0.0384 | 0.0352 | 0.0376 | 0.0002 | **0.0001** | 0.0001 |
| | $XB\backslash XB_{FSC}$ -std | 9.4e-006 | 2.34e-11 | 3.77e-13 | 9.97e-05 | **2.67e-05** | 4.87e-05 |
| NG20-T-2 (NG20-S2 ⇒ NG20-T2) | NMI-mean | 0.0201 | -- | 0.0282 | 0.8587 | -- | **0.9034** |
| | NMI-std | 0 | -- | 4.24e-18 | 0.0013 | -- | **1.24e-16** |
| | RI-mean | 0.4887 | -- | 0.4998 | 0.9574 | -- | **0.9750** |
| | RI-std | 6.20e-17 | -- | 0 | 0.0004 | -- | **1.24e-16** |
| | $XB\backslash XB_{FSC}$ -mean | 0.0422 | -- | 0.0362 | 0.1017 | -- | **0.0585** |
| | $XB\backslash XB_{FSC}$ -std | 2.81e-09 | -- | 5.95e-09 | 0.0071 | -- | **0.0026** |
| NG20-T1 (NG20-S3 ⇒ NG20-T1) | NMI-mean | 0.0184 | -- | 0.0251 | 0.5561 | -- | **0.6091** |
| | NMI-std | 0 | -- | 0.0002 | 0.0135 | -- | **0.0012** |
| | RI-mean | 0.2271 | -- | 0.3036 | 0.7573 | -- | **0.7934** |
| | RI-std | 0 | -- | 0.0004 | 0.0215 | -- | **0.0001** |
| | $XB\backslash XB_{FSC}$ -mean | 0.0384 | -- | 0.0361 | 0.0002 | -- | **0.0001** |
| | $XB\backslash XB_{FSC}$ -std | 9.4e-006 | -- | 0.0010 | 9.97e-05 | -- | **5.72e-05** |

* For a ⇒ b, b is the dataset in the target domain and a is the dataset in the corresponding source domain.

*C. 20 Newsgroups Text Clustering*

In this subsection, the proposed four TPFC algorithms are further experimented to evaluate their text clustering performance. The 20 newsgroups (20NG) text datasets [57] were adopted. In order to simulate the application scenarios studied in this paper, different categories of text data were adopted and each of them contains different sub-clusters in the source domain and the target domain with the details shown in Table XIV. In Table XV, four pairs of datasets with different source domains and target domains are presented for performance evaluation.

TABLE XIV
THE CLUSTERS OF 20 NEWSGROUPS TEXT DATA USED IN THIS STUDY.

| Cluster | Sub-cluster | | | | |
|---|---|---|---|---|---|
| *comp* | *comp. graphics* | *comp.os.ms -windows. misc* | *comp.sys. ibm.pc. hardware* | *comp. sys.mac. hardware* | *comp. windows.x* |
| *rec* | *rec. autos* | *rec. motorcycles* | *rec. sport. baseball* | *rec.sport. hockey* | - |
| *sci* | *sci.crypt* | *sci.electronics* | *sci.med* | *sci.space* | - |
| *talk* | *talk. politics. uns* | *talk. politics. mideast* | *talk. politics. misc-* | - | - |

In Table XV, four dataset pairs between the source and target domains are given. Of them, two dataset pairs (NG20-S1=>NG20-T1, NG20-S2=>NG20-T1) have the same numbers of clusters in both domains, while the other two dataset pairs (NG20-S2=>NG20-T2, NG20-S3=>NG20-T2) have different numbers of clusters in both domain. From Table XV, we can see that the constructed datasets have the following distinctive characteristics.

(1) The data in the source and target domains are related to some extent. The data in both domains belong to four different clusters, i.e., comp, rec, sci and talk.

(2) Although the data in both domains are related, there exist obvious differences between them. The same cluster in different domains contains very different sub-clusters. For example, in dataset pair NG20-S1=>NG20-T1 the source data of the rec cluster belong to three sub-clusters, i.e., rec.autos, rec.motorcycles and rec.sport.baseball, while the target data only belong to the sub-cluster rec.sport.hockey. Thus, the data distributions in both domains are related but different to some extent.

(3) The size of the dataset in the target domain is much smaller than that in the source domain. There is a practical requirement to improve the clustering effect on the data in the target domain by effectively using the knowledge of the source domain.

For the adopted data, dimensionality reduction has been applied by using the BOW toolkit [58] to effectively preprocess the high dimensional data into the final data containing 800 effective features used for clustering. For the proposed algorithms, the appropriate parameters of TFCM/E-TFCM and TFSC/E-TFSC have been determined with the XB and XBFSC indices, respectively.

In Table XVI, the clustering results on the datasets under the optimized parameters are reported, including the means and standard deviations of $NMI$, $RI$ and $XB / XB_{FSC}$ after 20 runs of every algorithm.

(1) In Table XVI, we see that TFCM and TFSC are not applicable for the datasets NG20-T1 (NG20-S2=>NG20-T2) and NG20-T1 (NG20-S3=>NG20-T1) since the numbers of clusters between the source domain and the target domain are different.

(2) As shown in Table XVI, in this high dimensional dataset, the three subspace clustering algorithms FSC, TFSC and E-TFSC have shown obvious advantage to the three non-subspace clustering algorithms FCM ,TFCM and E-TFCM. Again, the experimental results follow those in the previous experiments that the proposed four TPFC algorithms TFCM, E-TFCM, TFSC and E-TFSC are superior to their non-transfer learning counterparts.

(3) From Table XVI, we can see that for the datasets NG20-T1 (NG20-S1=>NG20-T1) and NG20-T1 (NG20-S2=>NG20-T1), where the numbers of clusters between the source and target domains are the same, TFCM is superior to E-TFCM and TFSC is superior to ETFSC. This is due to the fact that TFCM and TFSC have used more knowledge items for knowledge transfer in their objective functions than E-TFCM and E-FSC, respectively.

*D. Comparison with Related Clustering Algorithms*

In this subsection, the proposed four TPFC algorithms are finally compared with several related works, namely, two transfer clustering algorithms, i.e., self-taught clustering (STC) [16] and transfer spectral clustering (TSC) [17], two collaborative clustering algorithms CombKM and co-clustering (DRCC) [39] and a multi-task clustering algorithm LSSMTC [40] on both synthetic and real-world datasets used in subsections IV-B & C. For the proposed clustering algorithms, the appropriate parameters are determined with the XB or XBFSC indices. For the LSSMTC algorithm, the regularization parameter $\lambda \in [0,1]$ has been tried with the values in the set of $\{0.01,0.1,0.3,0.5,0.7,0.9,1\}$ and for the co-clustering algorithm, the related regularization parameters, $\lambda$ and $\mu$, have been set as $\lambda=\mu$ and tried in the set of $\{0.01, 0.1,0.3,0.5,0.7,0.9,1\}$. The parameter setting about TSC and STC is referred to [17]. Especially, for TSC algorithm it is required that the dimensional number of data is not smaller than the number of clusters. Thus, it is not applicable to two low dimensional datasets, i.e., T1, T2-1 and T2-2.

*1) Situation A: The Same Numbers of Clusters between the Source and Target Domains*

In this subsection, the clustering performance on the datasets with the same number of clusters between the source domain and the target domain is evaluated. For all the algorithms, the results obtained with the optimized parameter setting are reported in term of the means of 20 runs. Since $XB$ and $XB_{FSC}$ are not suitable to the no-fuzzy measure based methods, they are not reported for this type of algorithms. The experimental results are reported in Table XVII, based upon which the following observations can be made.

(1) The proposed TFCM, E-TFCM, TFSC and E-TFSC algorithms are highly competitive to the five related algorithms which have used the knowledge or data in the other tasks. In particular, TFSC and E-TFSC outperform the other algorithms in most of the adopted datasets.

(2) Compared TFSC with TFCM, it is obvious that the former outperforms the latter in the high dimensional datasets, such as T4 and two 20NG datasets.

In addition, the performance index $XB$ and modified $XB$ obtained by different fuzzy measure based algorithms are compared in Table XVIII. From the results, we can see that the proposed transfer fuzzy clustering algorithms have obtained improved clustering performance compared with the corresponding non-transfer counterparts.

Finally, the mean running time of 20 run of different algorithms on several datasets has been reported in Table XIX. From these results, we can see that the speed of the proposed transfer clustering algorithms is similar to that of the corresponding non-transfer counterparts, which also validates the theoretical analysis in subsection III-D.

TABLE XVII

PERFORMANCE COMPARISON OF THE BEST RESULTS OBTAINED BY TPFC AND THE RELATED ALGORITHMS ON DIFFERENT SYNTHETIC AND REAL-WORLD DATASETS BY USING NMI AND RI IN THE SCENE WITH THE SAME NUMBERS OF CLUSTERS BETWEEN THE SOURCE DOMAIN AND THE TARGET DOMAIN.

| Datasets | Indices | Algorithms | | | | | | | | |
|---|---|---|---|---|---|---|---|---|---|---|
| | | **LSSMTC** | **CombKM** | **DRCC** | **STC** | **TSC** | **TFCM** | **E-TFCM** | **TFSC** | **E-TFSC** |
| T1 (S1-1 $\Rightarrow$ T1)* | NMI-mean | 0.6972 | 0.6453 | 0.5811 | 0.7314 | - | 0.7850 | 0.7487 | **0.8751** | 0.8566 |
| | NMI-std | 1.35e-016 | 0 | 0.0201 | 0 | - | 1.17e-16 | 0 | **9.61e-17** | 0 |
| | RI-mean | 0.8807 | 0.8446 | 0.8318 | 0.9041 | - | 0.9187 | 0.9037 | **0.9651** | 0.9586 |
| | RI-std | 0 | 0 | 0.0113 | 0 | - | 0 | 0 | **0** | 0 |
| T1 (S1-2 $\Rightarrow$ T1) | NMI-mean | 0.6354 | 0.5301 | 0.5811 | 0.6455 | - | **0.6603** | 0.6593 | 0.6313 | 0.6307 |
| | NMI-std | 0 | 0 | 0.0201 | 0 | - | **0.0447** | 0.0490 | 0.0461 | 0.0454 |
| | RI-mean | 0.8491 | 0.7858 | 0.8318 | 0.7992 | - | **0.8605** | 0.8604 | 0.8459 | 0.8457 |
| | RI-std | 0 | 0 | 0.0113 | 0 | - | **0.0327** | 0.0306 | 0.0269 | 0.0271 |
| T4 (S4 $\Rightarrow$ T4) | NMI-mean | 0.5266 | 0.0561 | 0.4530 | 0.9382 | 0.9875 | 0.5536 | 0.5295 | **1** | 0.9878 |
| | NMI-std | 1.17e-016 | 7.31e-018 | 5.85e-017 | 0.0207 | 0.0023 | 7.31e-018 | 1.17e-16 | **1.17e-016** | 0.0271 |
| | RI-mean | 0.7463 | 0.4542 | 0.7056 | 0.9853 | 0.9927 | 0.7791 | 0.7535 | **1** | 0.9955 |
| | RI-std | 2.34e-016 | 5.85e-017 | 1.17e-016 | 0.0112 | 1.73e-03 | 1.17e-016 | 0 | **0** | 0.0098 |
| NG20-T1 (NG20-S1 $\Rightarrow$ NG20-T-1) | NMI-mean | 0.0407 | 0.0331 | 0.0284 | 0.1289 | 0.6315 | 0.0314 | 0.0247 | **0.6711** | 0.6394 |
| | NMI-std | 0 | 0 | 0.0011 | 0 | 0 | 0 | 0 | **0.0056** | 0.0157 |
| | RI-mean | 0.4468 | 0.3372 | 0.3371 | 0.3784 | 0.8489 | 0.3369 | 0.3178 | **0.8462** | 0.7972 |
| | RI-std | 0 | 0 | 5.22e-004 | 0 | 0 | 0 | 0 | **0.0088** | 0.0464 |
| NG20-T1 (NG20-S2 $\Rightarrow$ NG20-T1) | NMI-mean | 0.0383 | 0.0245 | 0.0284 | 0.1060 | 0.5674 | 0.0271 | 0.0196 | **0.6383** | 0.6270 |
| | NMI-std | 0 | 0.0013 | 0.0011 | 0 | 0 | 0 | 0 | **0.0139** | 0.0266 |
| | RI-mean | 0.4472 | 0.3367 | 0.3371 | 0.3728 | 0.8133 | 0.3273 | 0.2886 | **0.8065** | 0.8067 |
| | RI-std | 6.79e-017 | 7.72e-004 | 5.22e-004 | 0 | 0 | 0 | 0 | **0.0069** | 0.0172 |

* For a $\Rightarrow$ b, b is the dataset in the target domain and a is the dataset in the corresponding source domain.

TABLE XVIII
PERFORMANCE COMPARISON OF THE BEST RESULTS OF DIFFERENT FUZZY MEASURE BASED ALGORITHMS BY USING THE XB AND THE MODIFIED XB INDICES IN THE SCENE WITH THE SAME NUMBER OF CLUSTERS BETWEEN THE SOURCE DOMAIN AND THE TARGET DOMAIN.

| **Datasets** | | **Algorithms** | | | | | |
|---|---|---|---|---|---|---|---|
| | | **FCM** | **TFCM** | **E-TFCM** | **FSC** | **TFSC** | **E-TFSC** |
| | | ***XB*** | | | ***$XB_{FSC}$*** | | |
| T1 (S1-1 ⇒ T1)* | mean | 0.1911 | **0.1555** | 0.1801 | 5.2031 | **3.6361** | 4.0898 |
| | std | 0.0513 | **1.61e-15** | 0 | 0.0259 | **0.0001** | 0.0043 |
| T1 (S1-2 ⇒ T1) | mean | **0.1911** | 0.1928 | 0.1974 | 5.2123 | **5.2118** | 5.2121 |
| | std | **0.0513** | 0.0563 | 0.0612 | 0.0300 | **0.0305** | 0.0303 |
| T4 (S4 ⇒ T4) | mean | 2.04e+06 | **1.93e+05** | 2.40e+05 | 2.3418 | **0.4221** | 0.4785 |
| | std | 6.47e+06 | **5.10e+05** | 5.37e+05 | 0.0702 | **2.89e-05** | 0.0675 |
| NG20-T1 (NG20-S1 ⇒ NG20-T1) | mean | 0.0384 | **0.0341** | 0.0362 | 0.0002 | **1.53e-05** | 0.0001 |
| | std | 9.4e-006 | **8.0e-17** | 1.19e-08 | 9.97e-05 | **1.70e-06** | 2.97e-05 |
| NG20-T1 (NG20-S2 ⇒ NG20-T1) | mean | 0.0384 | **0.0352** | 0.0376 | 0.0002 | **0.0001** | 0.0001 |
| | std | 9.4e-006 | **2.34e-11** | 3.77e-13 | 9.97e-05 | **2.67e-05** | 4.87e-05 |

* For a ⇒ b, b is the dataset in the target domain and a is the dataset in the corresponding source domain.

TABLE XIX
RUNNING TIME (SECONDS) OF DIFFERENT ALGORITHMS IN THE SCENE WITH THE SAME NUMBERS OF CLUSTERS BETWEEN THE SOURCE DOMAIN AND THE TARGET DOMAIN.

| **Datasets** / **Algorithms** | T1 (S1-1 ⇒ T1)* | | T1 (S1-1 ⇒ T1) | | T4 (S4 ⇒ T4) | | NG-20-1 (NG20-S1 ⇒ NG20-T1) | | NG-20-2 (NG20-S1 ⇒ NG20-T1) | |
|---|---|---|---|---|---|---|---|---|---|---|
| | mean | std | mean | std | mean | std | mean | std | mean | std |
| **FCM** | 0.0149 | 0.0041 | 0.0149 | 0.0041 | 0.0465 | 0.0128 | 3.8649 | 0.0888 | 3.8649 | 0.0888 |
| **TFCM** | 0.0152 | 0.0018 | 0.0167 | 0.0055 | 0.0480 | 0.0076 | 12.2431 | 0.0465 | 12.9361 | 0.0727 |
| **E-TFCM** | 0.0342 | 0.0084 | 0.0264 | 0.0014 | 0.0497 | 0.0127 | 7.9117 | 0.4043 | 8.2090 | 0.0681 |
| **FSC** | 0.0174 | 0.0104 | 0.0174 | 0.0104 | 0.0156 | 0.0033 | 2.8332 | 0.7614 | 2.8332 | 0.7614 |
| **TFSC** | 0.0131 | 0.0009 | 0.0131 | 0.0009 | 0.0192 | 0.0072 | 4.7566 | 1.2696 | 4.9552 | 0.4510 |
| **E-TFSC** | 0.0185 | 0.0067 | 0.0177 | 0.0066 | 0.0135 | 0.0013 | 3.2889 | 0.3857 | 3.7263 | 0.3587 |
| **LSSMTC** | 0.0779 | 0.0033 | 0.0737 | 0.0038 | 0.7289 | 0.0062 | 24.9806 | 0.2003 | 24.9384 | 1.1288 |
| **CombKM** | 0.0036 | 2.3606e-004 | 0.0069 | 0.0033 | 0.0162 | 0.0026 | 2.9105 | 1.0718 | 2.7470 | 0.5320 |
| **DRCC** | 8.14e-004 | 5.52e-005 | 8.14e-004 | 5.52e-005 | 0.0041 | 0.0028 | 32.2058 | 2.1918 | 36.7134 | 5.3475 |
| **STC** | 0.7984 | 0.1847 | 0.0724 | 2.68e-005 | 43.317 | 1.251 | 368.2896 | 7.25e-005 | 401.4863 | 8.40e-005 |
| **TSC** | - | - | - | - | 5.0703 | 0.1515 | 492.0087 | 60.5947 | 511.7551 | 107.7256 |

* For a ⇒ b, b is the dataset in the target domain and a is the dataset in the corresponding source domain.

*2) Situation B: Different Numbers of Clusters between the Source and Target Domains*

In this subsection, the clustering performance are reported on the datasets in which the numbers of clusters between the source domain and the target domain are different. In this aspect, since most related algorithms are not applicable, their results are not provided. Finally, three available algorithms, i.e., E-TFCM, E-TFSC and CombKM, are adopted for performance evaluation.

For the adopted three algorithms, the results obtained with the optimized parameter setting are reported in Tables XX and XXI in term of the means of 20 runs. From these results we can get the following observations.

(1) The proposed E-TFCM and E-TFSC algorithms outperform the existing algorithm CombKM for the datasets considered here.

(2) Although both E-TFSC and E-TFCM have comparable performance for the low dimensional datasets, the former is better obviously than the latter for the high dimensional datasets.

(3) Table XXI shows that all three algorithms are comparable in the running time.

TABLE XXI RUNNING TIME (SECONDS) OF DIFFERENT ALGORITHMS IN THE SCENE WITH DIFFERENT NUMBERS OF CLUSTERS BETWEEN THE SOURCE DOMAIN AND THE TARGET DOMAIN.

| **Datasets** | **Indices** | **Algorithms** | | |
|---|---|---|---|---|
| | | E-TFCM | E-TFSC | CombKM |
| T2-1 (S2-1 ⇒ T2-1) * | time-mean | 0.0274 | 0.0075 | 0.0289 |
| | time-std | 0.0088 | 0.0015 | 0.0105 |
| T2-1 (S2-2 ⇒ T2-1) | time-mean | 0.0228 | 0.0187 | 0.0255 |
| | time-std | 0.0010 | 0.0075 | 0.0141 |
| T2-2 (S2-1 ⇒ T2-2) | time-mean | 0.0334 | 0.0225 | 0.0297 |
| | time-std | 0.0070 | 0.0165 | 0.0072 |
| T2-2 (S2-2 ⇒ T2-2) | time-mean | 0.0297 | 0.0269 | 0.0241 |
| | time-std | 0.0021 | 0.0122 | 0.0102 |
| T3-1 (S3-1 ⇒ T3-1) | time-mean | 0.0298 | 0.0066 | 0.0133 |
| | time-std | 0.0026 | 0.0006 | 0.0015 |
| T3-1 (S3-2 ⇒ T3-1) | time-mean | 0.0283 | 0.0098 | 0.0148 |
| | time-std | 0.0011 | 0.0016 | 0.0049 |
| T3-2 (S3-1 ⇒ T3-2) | time-mean | 0.0356 | 0.0094 | 0.0144 |
| | time-std | 0.0056 | 0.0022 | 0.0021 |
| T3-2 (S3-2 ⇒ T3-2) | time-mean | 0.0312 | 0.0082 | 0.0169 |
| | time-std | 0.0044 | 0.0014 | 0.0030 |
| NG20-T2 (NG20-S2 ⇒ NG20-T2) | time-mean | 3.3604 | 2.4864 | 8.1792 |
| | time-std | 0.0113 | 0.1299 | 0.0569 |
| NG20-T1 (NG20-S3 ⇒ NG20-T1) | time-mean | 5.8570 | 8.3636 | 10.1377 |
| | time-std | 0.0278 | 0.6391 | 0.2699 |

* For a ⇒ b, b is the dataset in the target domain and a is the dataset in the corresponding source domain.

TABLE XX
PERFORMANCE COMPARISON OF THE BEST RESULTS OBTAINED BY THE PROPOSED TPFC ALGORITHMS AND THE RELATED ALGORITHMS ON DIFFERENT DATASETS BY USING NMI AND RI IN THE SCENE WITH THE DIFFERENT NUMBERS OF CLUSTERS BETWEEN THE SOURCE DOMAIN AND THE TARGET DOMAIN.

| **Datasets** | **Indices** | **Algorithms** | | |
|---|---|---|---|---|
| | | E-TFCM | E-TFSC | CombKM |
| T2-1 (S2-1 ⟹ T2-1)* | NMI-mean | 0.7627 | **0.8015** | 0.5674 |
| | NMI-std | 0 | **1.24e-16** | 0.1002 |
| | RI-mean | 0.9158 | **0.9268** | 0.7836 |
| | RI-std | 0 | **0** | 0.0795 |
| T2-1 (S2-2 ⟹ T2-1) | NMI-mean | **0.6572** | 0.6254 | 0.5001 |
| | NMI-std | **0.0386** | 0.0475 | 0.0302 |
| | RI-mean | **0.8605** | 0.8415 | 0.7557 |
| | RI-std | **0.0258** | 0.0238 | 0.0278 |
| T2-2 (S2-1 ⟹ T2-2) | NMI-mean | 0.6612 | **0.7277** | 0.4249 |
| | NMI-std | 0 | **0.0061** | 0.0895 |
| | RI-mean | 0.8750 | **0.8947** | 0.7148 |
| | RI-std | 1.17e-16 | **0.0058** | 0.0682 |
| T2-2 (S2-2 ⟹ T2-2) | NMI-mean | 0.4870 | **0.5251** | 0.4000 |
| | NMI-std | 0.0617 | **0.0662** | 0.0547 |
| | RI-mean | 0.7641 | **0.7745** | 0.6843 |
| | RI-std | 0.0444 | **0.0414** | 0.0314 |
| T3-1 (S3-1 ⟹ T3-1) | NMI-mean | **0.8114** | **0.8114** | 0.6377 |
| | NMI-std | **0** | **1.24e-16** | 0.0590 |
| | RI-mean | **0.9344** | **0.9344** | 0.8342 |
| | RI-std | **0** | **0** | 0.0364 |
| T3-1 (S3-2 ⟹ T3-1) | NMI-mean | 0.7882 | **0.7991** | 0.6289 |
| | NMI-std | 1.17e-16 | **0.0276** | 0.0437 |
| | RI-mean | 0.9145 | **0.9282** | 0.8307 |
| | RI-std | 1.17e-16 | **0.0139** | 0.0265 |
| T3-2 (S3-1 ⟹ T3-2) | NMI-mean | **0.8142** | 0.7868 | 0.5788 |
| | NMI-std | **0** | 1.24e-16 | 0.2206 |
| | RI-mean | **0.9365** | 0.9220 | 0.8290 |
| | RI-std | **1.24e-16** | 0 | 0.0959 |
| T3-2 (S3-2 ⟹ T3-2) | NMI-mean | 0.7014 | **0.7076** | 0.4554 |
| | NMI-std | 0 | **0.0235** | 0.1570 |
| | RI-mean | 0.8773 | **0.8794** | 0.7794 |
| | RI-std | 0 | **0.0133** | 0.0668 |
| NG20-T2 (NG20-S2 ⟹ NG20-T2) | NMI-mean | 0.0282 | **0.9034** | 0.0064 |
| | NMI-std | 4.24e-18 | **1.24e-16** | 0 |
| | RI-mean | 0.4998 | **0.9750** | 0.2795 |
| | RI-std | 0 | **1.24e-16** | 6.79e-017 |
| NG20-T1 (NG20-S3 ⟹ NG20-T1) | NMI-mean | 0.0251 | **0.6091** | 0.0135 |
| | NMI-std | 0.0002 | **0.0012** | 0.0011 |
| | RI-mean | 0.3036 | **0.7934** | 0.2074 |
| | RI-std | 0.0004 | **0.0001** | 2.28e-004 |

* For a ⟹ b, b is the dataset in the target domain and a is the dataset in the corresponding source domain.

*E. Comparison of Different Initialization Strategies*

In this subsection, we present a performance comparison of the proposed algorithms TFCM and TFSC with two different initialization strategies, i.e., random initialization (Rand-Init) and source domain knowledge based initialization (Know-Init) for partition matrix $\mathbf{U}$. In this paper, we only report the results of TFCM to save space. Similar results can be obtained for the TFSC algorithm. For TFCM, the source knowledge based initialization is realized by using the update rule of membership function in FCM.

Here we have designed two synthetic datasets containing twenty-clusters in the source domain and the target domain, respectively. There exists a difference between the two datasets' distributions as shown in Fig 7. The corresponding clustering results are shown in Fig. 8 and Table XXII. In Fig. 8, the true cluster centers (V-true), the cluster centers obtained in the source domain (V-source) by FCM and the cluster centers obtained in the target domain (V-tfcm) by TFCM are denoted by "∇", "◊" and "O" respectively. For random initialization, the algorithm was run ten times using the given parameters and in the left column of Fig. 8 only the results in a certain run are shown. For the source knowledge based initialization, the algorithm was run only once using the given parameters since the clustering results are stable.

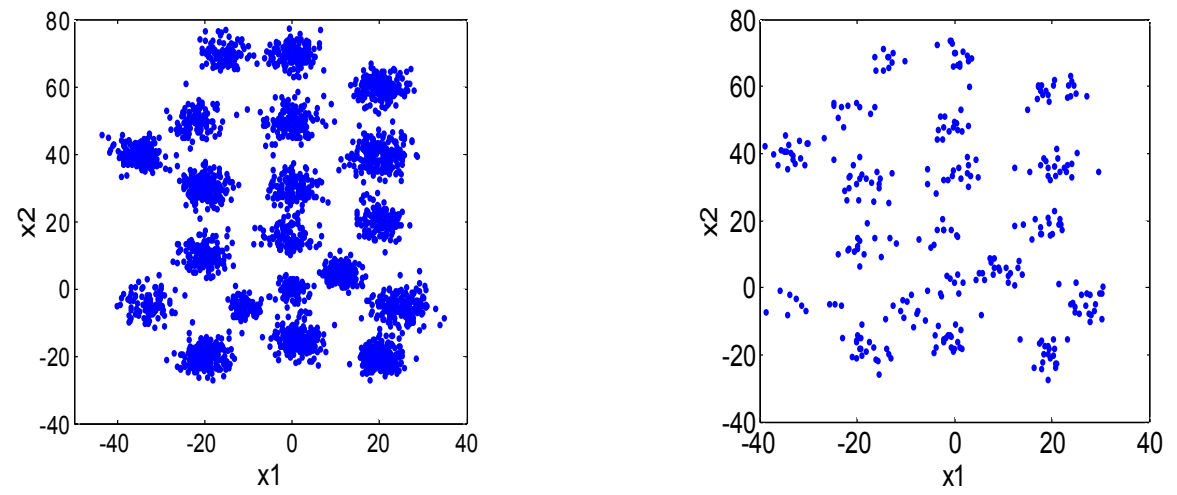

(a) (b)

Fig. 7 Two synthetic datasets containing 20 clusters: (a) data generated in the source domain S5; (b) data generated in the target domain T5.

From Fig. 8 and Table XXII, the following observations can be obtained.

(1) The left column of Fig. 8 and Table XXII show that with less source knowledge ($\lambda_1 = \lambda_2 = 0$ and $\lambda_1 = \lambda_2 = 0.005$), the random initialization strategy makes TFCM unstable and the imperfect matching between the clusters of both domains occurs, resulting in worse clustering results accordingly. However, with more source knowledge ($\lambda_1 = \lambda_2 = 0.1$ and $\lambda_1 = \lambda_2 = 1$), random initialization only slightly affects TFCM and more stable clustering results are obtained with better matching of the clusters between domains.

(2) The right column of Fig. 8 and Table XXII show that when the source knowledge based initialization strategy is adopted, a good matching of the clusters between both domains is always obtained. Thus, this strategy becomes promising in overcoming the challenge caused by potentially bad matching of clusters in the proposed algorithms.

(3) The experimental results in this subsection also show that the clustering performance of the proposed algorithms can be further improved by using a more appropriate initialization strategy, even though they with random initialization of the partition matrix have demonstrated better clustering performance than some related algorithms.

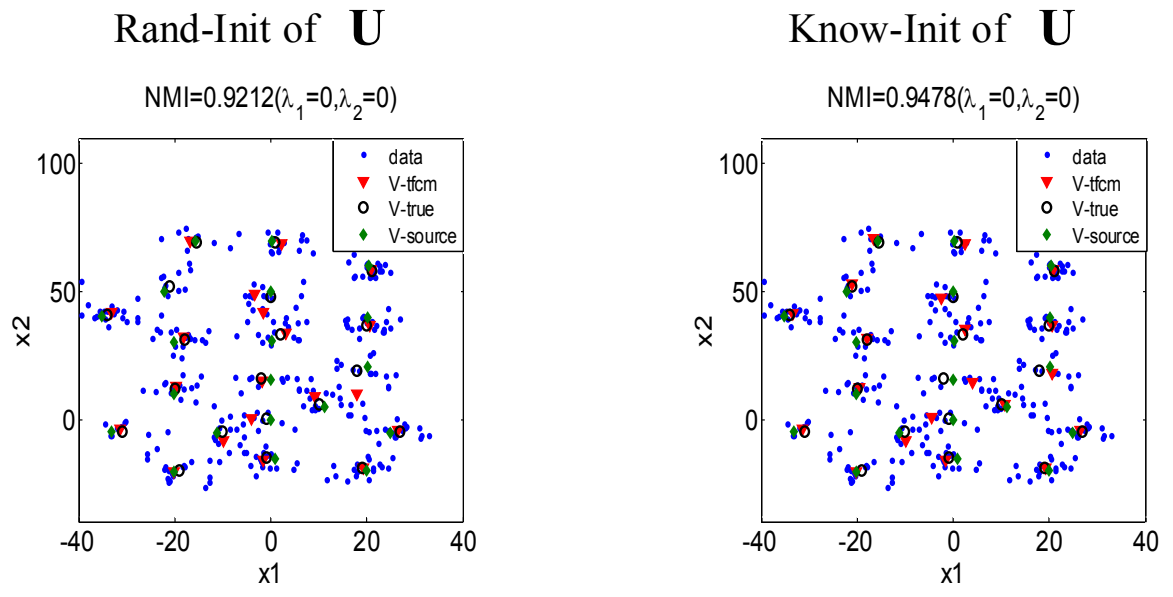

(a-1) (a-2)

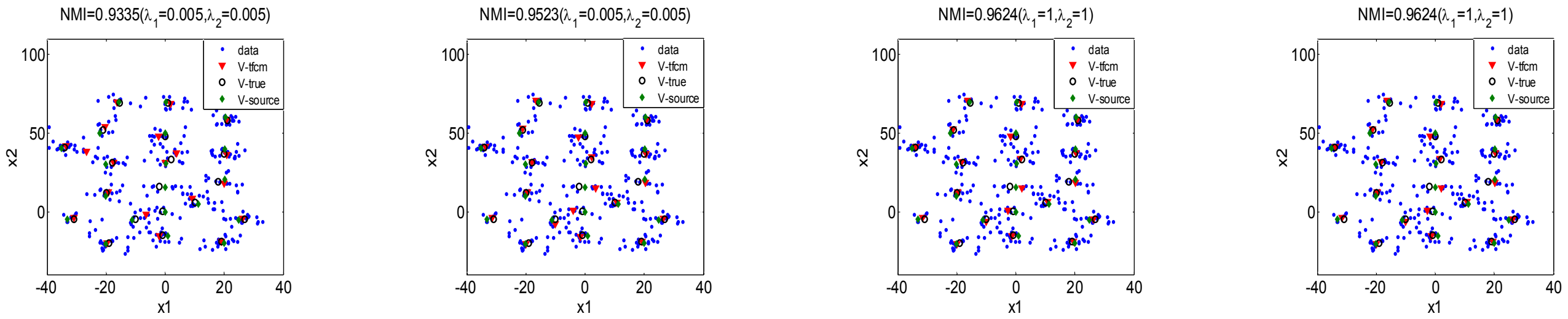

(b-1) (b-2) (d-1) (d-2)

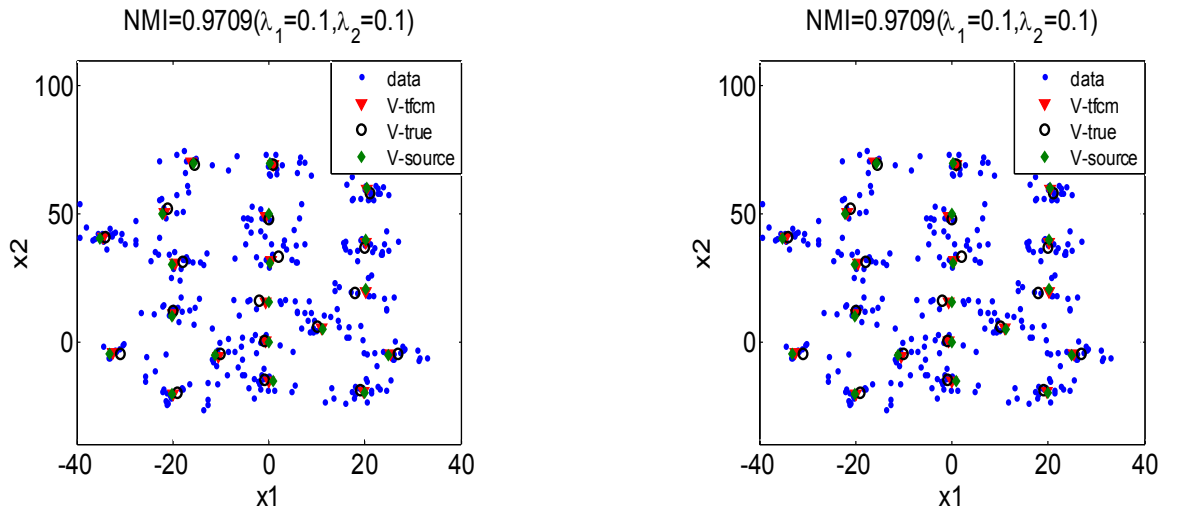

(c-1) (c-2)

Fig. 8 The clustering results of TFCM with two initialization strategies using different parameters. The left column shows the results obtained by the random initialization (Rand-Init) in a certain run and the right column depicts the results obtained by the source domain knowledge based initialization (Know-Init).

TABLE XXII
PERFORMANCE COMPARISON OF TFCM WITH TWO INITIALIZATION STRATEGIES USING DIFFERENT PARAMETERS ON THE SYNTHETIC DATASET CONTAINING TWENTY-CLUSTERS.

| | | Rand-Init of **U** (running 10 times) | | | Know-Init of **U** | | |
|---|---|---|---|---|---|---|---|
| | | NMI | RI | XB | NMI | RI | XB |
| $\lambda_1 = 0, \lambda_2 = 0$ | mean | 0.9224 | 0.9860 | 0.2585 | 0.9478 | 0.9921 | 0.1568 |
| | std | 0.0202 | 0.0041 | 0.0623 | 0 | 0 | 0 |
| $\lambda_1 = 0.005, \lambda_2 = 0.005$ | mean | 0.9403 | 0.9892 | 0.2002 | 0.9523 | 0.9928 | 0.1228 |
| | std | 0.0153 | 0.0034 | 0.0464 | 0 | 0 | 0 |
| $\lambda_1 = 0.1, \lambda_2 = 0.1$ | mean | 0.9709 | 0.9957 | 0.1038 | 0.9709 | 0.9957 | 0.1038 |
| | std | 1.17e-16 | 2.34e-16 | 6.63e-08 | 0 | 0 | 0 |
| $\lambda_1 = 1, \lambda_2 = 1$ | mean | 0.9624 | 0.9943 | 0.1205 | 0.9624 | 0.9943 | 0.1053 |
| | std | 2.34e-16 | 0 | 2.57e-17 | 0 | 0 | 0 |

### *F. Discussion*

In this subsection, let us further discuss an issue about the relations and limitations between the two dataset sizes (source/target) and the performance. It can be analyzed from three aspects. First, in this study, an assumption is given that the data in the target domain are scarce such that transfer learning is required to improve the clustering performance in the target domain. Therefore, when the size of the dataset in the target domain is small, the proposed transfer clustering algorithms can help to improve the clustering results when compared with the non-transfer counterparts. Second, when the size of the dataset in the target domain is much larger, the information is usually sufficient for obtaining the effective clustering by the traditional non-transfer clustering algorithms. In this case, even if the transfer clustering algorithms are adopted, the advantage will not be obvious. Third, in the proposed transfer clustering algorithms, the original raw data in the source domain are not used and only some knowledge, such as cluster centers, are adopted in the clustering procedure in the target domain. Thus, the reliability of the knowledge obtained in the source domain has an important influence on the clustering performance of the proposed transfer clustering algorithms. If the size of dataset in the source domain is small, the induced knowledge may not be reliable. When the size of the dataset in the source domain is large, more reliable knowledge can be expected from the source domain. With such an analysis, we can infer that the larger the size of the dataset in the source domain, the more reliable the knowledge obtained in the source domain and the more effective the proposed transfer clustering algorithms will be in the target domain.

## V. CONCLUSIONS

In this study, the concept of knowledge leveraging is used to develop a set of transfer prototype-based fuzzy clustering methods for the application scenarios where the data are limited or insufficiently available for effective clustering. Based on the introduced knowledge-leverage transfer learning mechanism, several transfer prototype-based fuzzy clustering algorithms are exploited to address the deficiency caused by the scarceness of data in the target domain. The proposed prototype-based fuzzy clustering algorithms learn from not only the data of the target domain but also the knowledge of the source domain. The experimental results have demonstrated the attractiveness and effectiveness of the proposed transfer prototype-based fuzzy clustering algorithms when compared with the existing classical prototype-based fuzzy clustering algorithms without the transfer learning abilities, and other related algorithms such as the multi-task clustering algorithm.

Although the proposed transfer prototype based fuzzy clustering algorithms have demonstrated their promising performance, more works can be further addressed about this research topic. One important work is how to incorporate more advantageous transfer mechanisms to develop better transfer prototype based clustering algorithms. Beside the adopted FCM and FSC methods, it is also very valuable to develop the corresponding transfer learning versions based on other prototype based clustering algorithms. In future, we will address these issues in depth.

## REFERENCES

[1] S. J. Pan and Q. Yang, "A survey on transfer learning," *IEEE Trans. Knowledge Data Engineering,* vol. 22, no. 10, pp. 1345–1359, 2010.

[2] J.W. Tao, F.L. Chung, and S.T. Wang, "On minimum distribution discrepancy support vector machine for domain adaptation," *Pattern Recognition*, vol. 45, no. 11, pp. 3962-3984, 2012.

[3] Z. Sun, Y.Q. Chen, J. Qi, and J.F. Liu, "Adaptive localization through transfer learning in indoor Wi-Fi environment," *Proc. 7th International Conference on Machine Learning and Applications*, pp. 331–336, 2008.

[4] S. Bickel, M. Brückner, and T. Scheffer, "Discriminative learning for differing training and test distributions," *Proc. 24th Int. Conf. Machine Learning,* pp. 81-88, 2007.

[5] N. D. Lawrence and J. C. Platt, "Learning to learn with the informative vector machine," *Proc. 21st Int. Conf. Machine Learning*, 2004.

[6] J. Gao, W. Fan, J. Jiang, and J. Han, "Knowledge transfer via multiple model local structure mapping," *Proc. 14th ACM SIGKDD Int. Conf. Knowledge Discovery and Data Mining*, pp. 283-291, 2008.

[7] L. Mihalkova and R. J. Mooney, "Transfer learning by mapping with minimal target data," *Proc. Assoc. for the Advancement of Artificial Intelligence (AAAI '08) Workshop Transfer Learning for Complex Tasks*, 2008.

[8] J. Davis and P. Domingos, "Deep transfer via second-order Markov logic," *Proc. Assoc. for the Advancement of Artificial Intelligence (AAAI '08) Workshop Transfer Learning for Complex Tasks*, 2008.

[9] S.J. Pan, I. W. Tsang, J. T. Kwok, and Q. Yang, "Domain adaptation via transfer component analysis," *IEEE Trans. Neural Networks*, vol. 22, no.2, pp.199-210, 2011.

[10] Z. Wang, Y. Song, and C. Zhang, "Transferred dimensionality reduction," *Proc. European Conf. Machine Learning and Knowledge Discovery in Databases (ECML/PKDD '08)*, pp. 550-565, 2008.

*[11]* P. Yang, Q. Tan, and Y. Ding, "Bayesian task-level transfer learning for non-linear regression," *Proc. Int. Conf. on Computer Science and Software Engineering,* pp. 62-65, 2008.

[12] L. Borzemski and G. Starczewski, "Application of transfer regression to TCP throughput prediction," *Proc. First Asian Conference on Intelligent Information and Database Systems*, pp. 28-33, 2009.

[13] J. Liu, Y. Chen, and Y. Zhang, "Transfer regression model for indoor 3D location estimation," *Lecture Note on Computer Science 5916,* pp. 603–613, 2010.

[14] Z.H. Deng, Y.Z. Jiang, K.S. Choi, F.L. Chung, and S.T. Wang, "Knowledge-Leverage based TSK fuzzy system modeling," *IEEE Trans. Neural Networks and Learning Systems*, vol. 24, no. 8, pp. 1200-1212, 2013.

[15] Z.H. Deng, Y.Z. Jiang, F.L. Chung, H. Ishibuchi, and S.T. Wang, "Knowledge-Leverage based fuzzy system and its modeling," *IEEE Trans. Fuzzy systems,* vol.21, no. 4, pp. 597-609, 2013.

[16] W. Dai, Q. Yang, G. Xue, and Y. Yu, "Self-taught clustering," *Proc. 25th Int. Conf. Machine* Learning, pp. 200-207, 2008.

[17] W.H. Jiang and F.L. Chung, "Transfer spectral clustering," *In Machine Learning and Knowledge Discovery in Databases*, pp. 789-803, 2012.

[18] J. Dunn, "A fuzzy relative of the ISODATA process and its use in detecting compact well separated clusters," *Journal of Cybernetics*, vol. 3, no. 3, pp.32–57, 1973.

[19] J. C. Bezdek, *Pattern Recognition with Fuzzy Objective Function Algorithms*. New York Plenum, 1981.

[20] R. Krishnapuram and J. M. Keller, "A possibilistic approach to clustering." *IEEE Trans. on Fuzzy Systems*, vol.1, no. 2, pp. 98-110, 1993.

[21] R.P. Li and M.A. Mukaidono, "A maximum entropy approach to fuzzy clustering," *Proc on IEEE Int Conf Fuzzy Syst.* Yokohama, Japan, pp. 2227-2232, 1995.

[22] J.Z. Huang, M.K. Ng, H. Rong, and Z. Li, "Automated variable weighting in k-means type clustering," *IEEE Trans. Pattern Anal. Mach. Intell.* 27 (5) (2005) 657–668.

[23] P. S. Bradley, and Q. L. Mangasarian, "K-plane clustering," *Journal of Global Optimization*, vol. 16, no. 1, pp. 23-32, 2000.

[24] M. Ester, H.P. Kriegel, J. Sander, and X. Xu, "A density-based algorithm for discovering clusters in large spatial databases with noise," *Proc. 2nd International Conference on KDD*, pp. 226-231, 1996.

[25] J. Sander, M. Ester, H.P. Kriegel, and X. Xu, "Density-based clustering in spatial databases: The algorithm GDBSCAN and its applications," *Data Mining and Knowledge Discovery*, vol. 2, no. 2, pp. 169-194, 1998.

[26] A. Hinneburg and D.A. Keim, "An efficient approach to clustering in large multimedia databases with noise," *Proc. 2nd International Conference on KDD*, pp. 58–65, 1998.

[27] M. Ankerst, M.M. Breunig, H.P. Kriegel, and J. Sander, "OPTICS: Ordering points to identify the clustering structure," *Proc. International Conference on Management of Data*, pp. 49–60, 1999.

[28] T. Pei, A. Jasra, D.J. Hand, A.X. Zhu, and C. Zhou, "DECODE: a new method for discovering clusters of different densities in spatial data," *Data Mining and Knowledge Discovery*, vol. 18, no. 3, pp. 337–369, 2009.

[29] J. Shi and J. Malik, "Normalized cuts and image segmentation," *IEEE Trans. Pattern Analysis and Machine. Intelligence*, vol. 22, no. 8, pp. 888–905, 2000.

[30] E. Hartuv and R. Shamir, "A clustering algorithm based on graph connectivity," *Information Processing Letter*, vol. 76, no. 4, pp. 175–181, 2000.

[31] A.Y. Ng, M.I. Jordan, and Y. Weiss, "On spectral clustering: Analysis and an algorithm," *Advances in Neural Information Processing Systems*, vol. 2, pp. 849-856, 2002.

[32] P.J. Qian, F.L. Chung, S.T. Wang, and Z.H. Deng, "Fast graph-based relaxed clustering for large data sets using minimal enclosing ball," *IEEE Trans. Systems, Man, and Cybernetics, Part B- Cybernetics*, vol. 42, no. 3, pp. 672-687, 2012.

[33] A. Ben-Hur, D. Horn, H. Siegelmann, and V. Vapnik, "Support vector clustering," *Journal of Machine Learning Research*, vol. 2, pp. 125–137, 2001.

[34] L. Tseng and S. Yang, "A genetic approach to the automatic clustering problem," *Pattern Recognition*, vol. 34, no. 2, pp. 415–424, 2001.

[35] T. Kohonen, "The self-organizing map," *Proc. IEEE*, vol. 78, no. 9, pp.1464–1480, 1990.

[36] Q.Q. Gu, and J. Zhou, "Co-clustering on manifolds," *Proc. 15th International Conference on KDD,* pp. 359-368, 2009.

[37] W. Pedrycz, "Collaborative fuzzy clustering," *Pattern Recognition Letters*, vol. 23, no. 14, pp. 1675-1686, 2002.

[38] W. Pedrycz, and P. Rai, "Collaborative clustering with the use of Fuzzy C-Means and its quantification," *Fuzzy Sets and Systems*, vol. 159, no. 18, pp. 2399-2427, 2008.

[39] Q.Q. Gu and J. Zhou, "Learning the shared subspace for multi-task clustering and transductive transfer classification," *Proc. 9th IEEE International Conference on Data Mining*, pp. 159-168, 2009.

[40] Z.H. Zhang, and J. Zhou, "Multi-task clustering via domain adaptation," *Pattern Recognition*, vol. 45, no. 1, pp. 465-473, 2012.

[41] M. Bilenko, S. Basu, and R.J. Mooney, "Integrating constraints and metric learning in semi-supervised clustering," *Proc. 21th International Conference on Machine learning*, pp.81 -88, 2004.

[42] V. Loia, W. Pedrycz, and S. Senatore, "Semantic web content analysis: A study in proximity-based collaborative clustering," *IEEE Trans. Fuzzy Systems*, vol. 15, no. 6, pp. 1294–1312, 2007.

[43] W. Pedrycz, "Fuzzy clustering with a knowledge-based guidance," *Pattern Recognition Letters*, vol. 25, no. 4, pp. 469–480, 2004.

[44] W. Pedrycz, V. Loia, and S. Senatore, "P-FCM: A proximity based fuzzy clustering," *Fuzzy Sets and Systems*, vol. 148, no. 1, pp. 21–41, 2004.

[45] F.L. Chung, S.T. Wang, Z.H. Deng, S.Chen, and D.W. Hu, "Clustering analysis of gene expression data based on semi-supervised visual clustering algorithm," *Soft Comput.* vol. 10, no. 11, pp. 981-993, 2006.

[46] M. Tabatabaei-Pour, K. Salahshoor, and B. Moshiri. "A modified k-plane clustering algorithm for identification of hybrid systems," *Proc. 6th World Congress on Intelligent Control and Automation*, pp. 1333-1337, 2006.

[47] Y. Wang, S.C Chen, D.Q. Zhang, and X.B. Yang, "Fuzzy k-Plane clustering algorithm," *Pattern Recognition and Artificial Intelligence*, vol. 20, no. 5, pp. 704-710, 2007.

[48] J. Liu and T.D. Pham, "A spatially constrained fuzzy hyper-prototype clustering algorithm," *Pattern Recognition*, vol. 45, no. 4, pp. 1759-1771, 2012.

[49] A Banerjee, I.S. Dhillon, J. Ghosh, et al, "Generative model based clustering of directional data," *Conf erence on Knowledge Discovery in Data*, Washington, DC, 2003.

[50] I.S. Dhillon and D.S.Modha1, "Concept decompositions for large sparse text data using clustering," *Machine Learning*, vol. 42 no.1, pp.143- 175, 2001.

[51] G.J. Gan and J.H. Wu, "A convergence theorem for the fuzzy subspace clustering (FSC) algorithm," *Pattern Recognition*, vol. 41, no. 6, pp. 1939–1947, 2008.

[52] L.P. Jing, M.K. Ng, and Z.X. Huang, "An entropy weighting k-means algorithm for subspace clustering of high-dimensional sparse data," *IEEE Trans. Knowledge and Data Engineering*, vol. 19, no. 8, pp. 1026–1041, 2007.

[53] Z.H. Deng, K.S. Choi, F.L. Chung, and S.T. Wang, "Enhanced soft subspace clustering integrating within-cluster and between-cluster information," *Pattern Recognition*, vol. 43, no. 3, pp. 767-781, 2010.

[54] R. Hathaway, J. Bezdek, and W. Tucker, "An improved convergence theorem for the fuzzy c-means clustering algorithms," in: J. Bezdek (Ed.), *Analysis of Fuzzy Information*, vol. III, CRC Press, Boca Raton, pp. 123–131, 1987.

[55] W. Zangwill, *Non-linear Programming: A Unified Approach*, Prentice-Hall, Englewood Cliffs, NJ, 1969.

[56] J. Liu, J. Mohammed, J. Carter, et al., "Distance-based clustering of CGH data," *Bioinformatics*, vol. 22, no. 16, pp. 1971–1978, 2006.

[57] W.Y. Dai, G.R. Xue, Q. Yang, and Y. Yu, "Co-clustering based classification for out-of-domain documents," *Proc. 13th International Conference on KDD*, pp. 210-219, 2007.

[58] A. K. McCallum, "Bow: A toolkit for statistical language modeling, text retrieval, classification and clustering," http://www-2.cs.cmu.edu/~mccallum/bow/.

[59] C. Domeniconi and M. Al-Razgan, "Weighted cluster ensembles: Methods and analysis," *ACM Transactions on Knowledge Discovery from Data (TKDD)*, vo. 2, no. 4, pp.1-40, 2009.

[60] G. Francesco, C. Domeniconi, and A. Tagarelli, "Projective clustering ensembles," Data Mining and Knowledge Discovery, vol. 26, no. 3, pp. 452-511, 2013.

[61] Z. H. Deng, K. S. Choi, J. Wang, S. T. Wang, "A survey on soft subspace clustering," CoRR abs/1409.5616, 2014.

[62] J. Yu and M. S. Yang, "Optimality test for generalized FCM and its application to parameter selection," *IEEE Trans. on Fuzzy Systems*, vol. 13, no. 1, pp. 164-176, 2005.

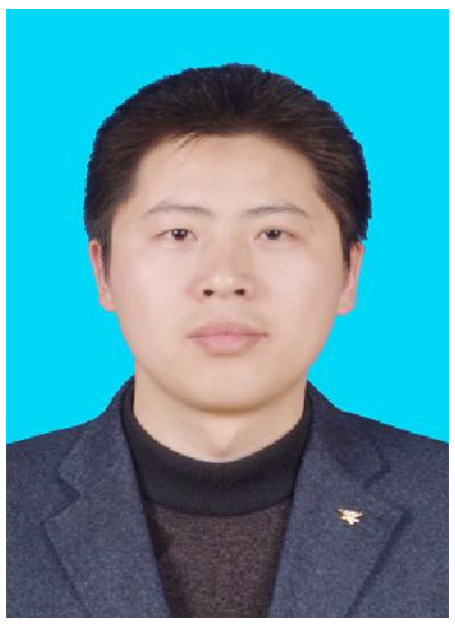

**Zhaohong Deng** (M'2012, SM'2014) received the B.S. degree in physics from Fuyang Normal College, Fuyang, China, in 2002, and the Ph.D. degree in light industry information technology and engineering from Jiangnan University, Wuxi, China, in 2008. He is currently a professor in the School of Digital Media, Jiangnan University and a visiting associate researcher in the University of California, Davis, CA, USA. His current research interests include computational intelligence and pattern recognition. He serves as the associate editor of four international editors including Neurocomputing, PLOS one. He has published about 80 papers in international/national journals.

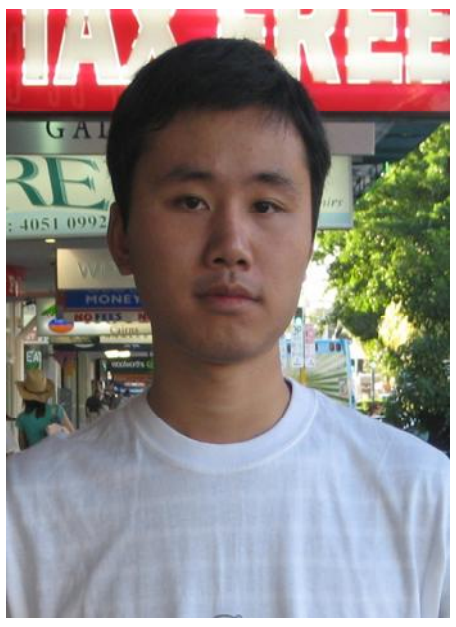

**Yizhang Jiang** (M'2012) is a Ph.D. candidate in the School of Digital Media, Jiangnan University. He has also been a research assistant in the Department of Computing, Hong Kong Polytechnic University, for two years. His research interests include pattern recognition, intelligent computation and their applications. He is the author/co-author of more than 20 research papers in international/national journals, including IEEE Trans. on Fuzzy Systems, IEEE Trans. on Neural Networks and Learning Systems, IEEE Trans. on Cybernetics, and Information Sciences. He has served as a reviewer or co-reviewer of several international conferences and journals, such as ICDM, TKDE, TFS, TNNLS, Pattern Recognition, Neurocomputing, and Neural Computing & Applications.

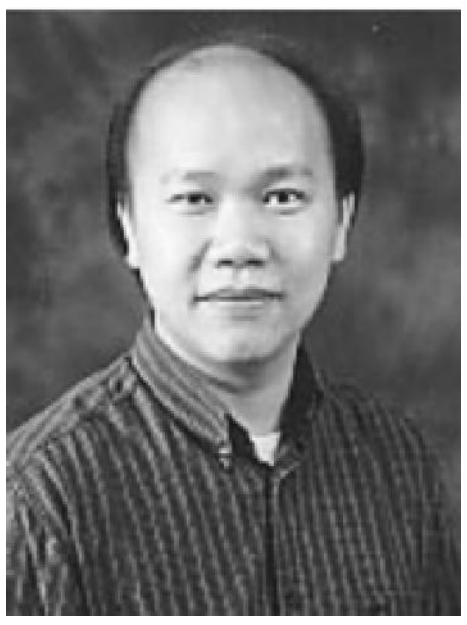

**Fu-lai Chung** (M'95) received the B.Sc. degree from the University of Manitoba, Winnipeg, MB, Canada, in 1987 and the M.Phil. and Ph.D. degrees from the Chinese University of Hong Kong, Hong Kong, in 1991 and 1995, respectively. In 1994, he joined the Department of Computing, Hong Kong Polytechnic University, where he is currently an Associate Professor. He has authored or coauthored over 80 journal papers published in the areas of soft computing, data mining, machine intelligence, and multimedia. His current research interests include transfer learning, social network analysis and mining, kernel learning, dimensionality reduction, and big data learning.

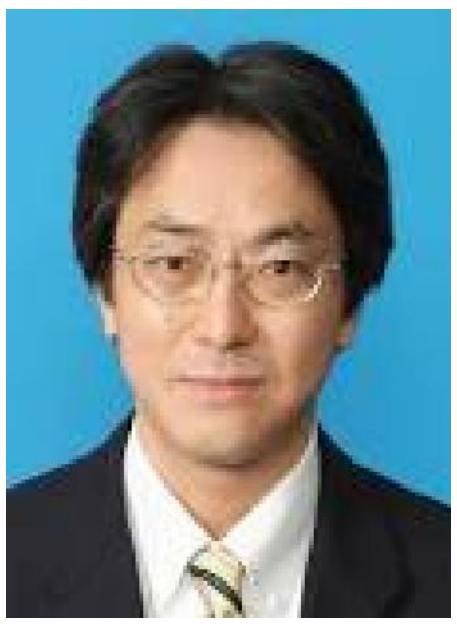

**Hisao Ishibuchi** (M'93–F'14) received the B.S., M.S. and Ph.D. in industrial engineering from Osaka Prefecture University, Osaka, Japan. Since 1999, he has been a Full Professor with Osaka Prefecture University. His research interests include artificial intelligence, neural fuzzy systems, and data mining. Dr. Ishibuchi is on the editorial boards of several journals, including the IEEE Transactions on Fuzzy Systems and the IEEE Transactions on Systems, Man, and Cybernetics—Part B.

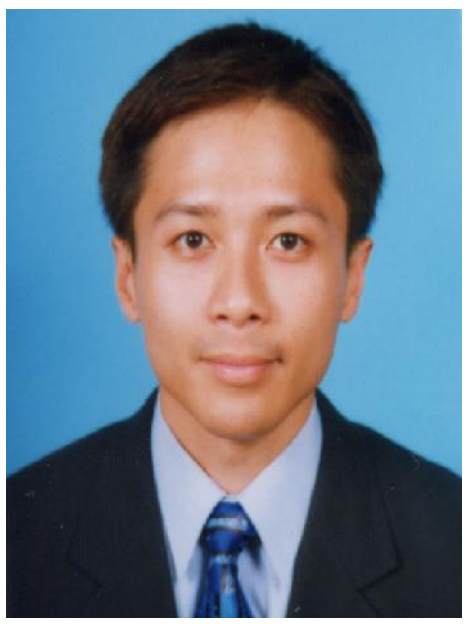

**Kup-Sze Choi** (M'97) received his Ph.D. degree in computer science and engineering from the Chinese University of Hong Kong. He is currently an associate professor at the School of Nursing, the Hong Kong Polytechnic University, and the leader of the Technology in Health Care research team. His research interests include computational intelligence, computer graphics, virtual reality, physics-based simulation, and their applications in medicine and health care.

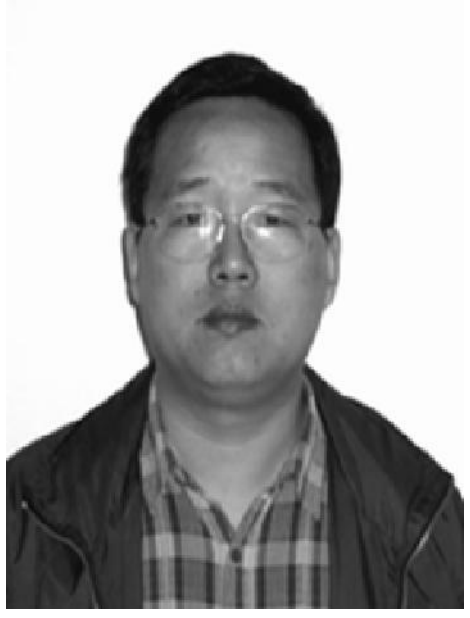

**Shitong Wang** received the M.S. degree in computer science from Nanjing University of Aeronautics and Astronautics, China, in 1987. He visited London University and Bristol University in U.K., Hiroshima International University in Japan, Hong Kong University of Science and Technology, Hong Kong Polytechnic University, as a Research Scientist, for over six years. Currently, he is a Full Professor of the School of Digital Media, Jiangnan University, China. His research interests include artificial intelligence, neuro-fuzzy systems, pattern recognition, and image processing. He has published about 80 papers in international/national journals and has authored seven books.